\documentclass{article}

\usepackage[preprint, nonatbib]{neurips_2025}

\usepackage[utf8]{inputenc} 
\usepackage[T1]{fontenc}    
\usepackage{hyperref}       
\usepackage{url}            
\usepackage{booktabs}       
\usepackage{amsfonts}       
\usepackage{nicefrac}       
\usepackage{microtype}      
\usepackage{xcolor}         
\usepackage{amsmath}
\usepackage{algorithm}
\usepackage{algorithmic}
\usepackage{graphicx}
\usepackage{bm}
\usepackage{tcolorbox} 
\usepackage{multirow}
\usepackage{float}

\title{Humble your Overconfident Networks: Unlearning Overfitting via Sequential Monte Carlo Tempered Deep Ensembles}

\author{
  Andrew Millard \thanks{Department of Electrical Engineering and Electronics, University of Liverpool. Contact: \texttt{firstname.lastname@liverpool.ac.uk}} \\
  \And
  Zheng Zhao \thanks{Department of Computer and Information Science, Link\"oping University. Contact:\texttt{firstname.lastname@liu.se}} \\
  \And
  Joshua Murphy \footnotemark[1] \\
  \And
  Simon Maskell \footnotemark[1]\\
}

\begin{document}

\maketitle

\begin{abstract}
    Sequential Monte Carlo (SMC) methods offer a principled approach to Bayesian uncertainty quantification but are traditionally limited by the need for full-batch gradient evaluations. We introduce a scalable variant by incorporating Stochastic Gradient Hamiltonian Monte Carlo (SGHMC) proposals into SMC, enabling efficient mini-batch-based sampling. Our resulting SMCSGHMC algorithm outperforms standard stochastic gradient descent (SGD) and deep ensembles across image classification, out-of-distribution (OOD) detection, and transfer learning tasks. We further show that SMCSGHMC mitigates overfitting and improves calibration, providing a flexible, scalable pathway for converting pretrained neural networks into well-calibrated Bayesian models.
\end{abstract}

\section{Introduction}
Sequential Monte Carlo (SMC) samplers \cite{DelMoral2006} are a class of Bayesian inference algorithms designed to approximate posterior distributions. While effective, they can be computationally expensive in high-dimensional settings, as they typically require full-batch evaluations. To address this limitation, scalable Bayesian inference methods such as Stochastic Gradient Langevin Dynamics (SGLD) \cite{welling2011bayesian} and Stochastic Gradient Hamiltonian Monte Carlo (SGHMC) \cite{chen2014stochasticgradienthamiltonianmonte} have been developed. These methods approximate gradients using mini-batches, a technique inspired by stochastic gradient descent (SGD) \cite{ruder2017overviewgradientdescentoptimization}, making them well-suited to large datasets and complex models like neural networks. By combining mini-batch gradient estimates with Langevin or Hamiltonian dynamics, SGLD and SGHMC can effectively explore the high-dimensional, non-convex landscapes typical of deep learning models.

Despite their scalability, SGLD and SGHMC require careful tuning of additional hyperparameters. SGHMC, in particular, introduces a noise model $\hat{V}$ to estimate gradient variance arising from the use of mini-batches, as well as a friction term $C$ to counteract the additional stochasticity. Furthermore, optimal performance often depends on designing a decreasing step size schedule, as discussed in \cite{zhang2020cyclicalstochasticgradientmcmc}. These requirements add complexity to the sampling process and can hinder practical implementation.

Methods such as Stochastic Weight Averaging Gaussian (SWAG) \cite{maddox2019simplebaselinebayesianuncertainty} and Laplacian Approximations (LA) \cite{daxberger2022laplacereduxeffortless, ritter2018scalable} offer further computationally efficient alternatives. These approaches are typically applied as post-processing steps to pretrained models, avoiding the need for retraining while still yielding uncertainty estimates. However, both SWAG and LA assume that the posterior can be well-approximated by a Gaussian distribution—an assumption that does not always hold in practice. 
This has been recently evidenced by \cite{izmailov2021bayesianneuralnetworkposteriors} showing that neural network posteriors are often highly non-Gaussian and multimodal.

To provide more accurate uncertainty estimates without the burden of extensive hyperparameter tuning, we propose a non-parametric, post-training sampling method that draws inspiration from both SGLD/SGHMC and SWAG/LA. From the former, we borrow the idea of using gradient-based dynamics—specifically, a simplified version of SGHMC that omits both added noise and friction—as a proposal mechanism. From the latter, we adopt the post-training philosophy: our method is applied after the network has been trained, avoiding the need to retrain or modify the training procedure. Since SMC relies on sequential importance sampling (SIS), it permits flexible proposal distributions as long as they satisfy certain conditions. This flexibility allows us to bypass the tuning of noise models, friction terms, and step sizes, while still drawing high-quality samples from the posterior.

The remainder of the paper is organized as follows.  
Section~\ref{sec:smc} provides an overview of Sequential Monte Carlo (SMC) samplers.  
Section~\ref{sec:hmc} discusses Hamiltonian Monte Carlo (HMC) dynamics and their basic stochastic variants, and describes how they can be incorporated within an SMC framework for scalable Bayesian inference.  
Section~\ref{sec:scalable} outlines how the resulting algorithm can be applied to Bayesian neural networks (BNNs) in large-scale deep learning settings.  
Sections~\ref{sec:experiments} and~\ref{sec:results} present the experimental setup and a discussion of results, respectively.

The main contributions of this work are threefold:

\begin{enumerate}
    \item We propose the use of an SGHMC-based proposal mechanism within an SMC sampler, enabling scalable Bayesian inference for high-dimensional neural networks.
    \item We demonstrate that our method can effectively mitigate overfitting, a common issue with SGD-based training, leading to improved expected calibration error (ECE) metrics.
    \item We show that improved calibration translates to stronger performance on several out-of-distribution (OOD) detection tasks and transfer learning benchmarks, outperforming standard Bayesian deep learning methods.
\end{enumerate}

\section{Sequential Monte Carlo Samplers} \label{sec:smc}
Sequential Monte Carlo (SMC) samplers extend importance sampling (IS) by iteratively adapting a set of weighted samples toward a target distribution $\pi(\bm{\theta})$.  
In IS, samples $\bm{\theta}$ are drawn from a proposal distribution $q(\bm{\theta})$ and reweighted to approximate expectations under $\pi(\bm{\theta})$; a pedagogical overview of Monte Carlo (MC) and IS is provided in Appendix~\ref{app:mc_tutorial} and Appendix~\ref{app:is_tutorial}.

A single IS iteration often fails when $q$ and $\pi$ differ significantly, leading to sample degeneracy. Sequential Importance Sampling (SIS)~\cite{maceachern1999sequential} addresses this by updating samples and weights iteratively. In a Bayesian framework, the posterior is given by:
\begin{equation}
p(\bm{\theta} \mid \mathcal{D}) \propto p(\mathcal{D} \mid \bm{\theta}) p(\bm{\theta}),
\end{equation}

where $p(\bm{\theta})$ (also often written as $q_0(\bm{\theta})$) is the prior and $p(\mathcal{D} \mid \bm{\theta})$ is the likelihood (often modeled by a neural network, such as a categorical distribution for classification). We propogate the samples using a Markov kernel $q_t(\bm{\theta}_t | \bm{\theta}_{t-1})$ and then weight them according to the SIS weight update:
\begin{equation}
    \mathbf{w}_t(\bm{\theta}_{1:t}) = \mathbf{w}_{t-1}(\bm{\theta}_{1:t-1}) p(\mathcal{D} \mid \bm{\theta}_t).
\end{equation}

This recursive update~\cite{chopin2020introduction} adjusts particle weights based on how well new proposals explain the data.

\begin{tcolorbox}[colback=gray!10, colframe=black!50, title=Remark: Degeneracy in SIS]
SIS suffers from sample degeneracy: most particles carry negligible weight after a few iterations. Resampling steps are introduced to address this in SMC.
\end{tcolorbox}

To mitigate degeneracy, SMC methods~\cite{delmoral2002sequential} incorporate resampling~\cite{douc2005comparison}, selectively amplifying high-weight particles while discarding low-weight ones. We employ an SMC framework similar to that in \cite{johansen2008particle}. The overall SMC procedure consists of:

\begin{enumerate}
    \item \textbf{Initialization:} Draw $J$ particles from the prior $\{\bm{\theta}_0^{(j)}\}_{j=1}^J \sim q_0(\bm{\theta})$, and set weights
    \begin{equation}
    \mathbf{w}_0^{(j)} = \frac{\pi(\bm{\theta}_0^{(j)})}{q_0(\bm{\theta}_0^{(j)})}.
    \label{equ:init_sample}
    \end{equation}

    \item \textbf{Iterative Updates (for $t=1$ to $T$):}
    \begin{enumerate}
        \item Normalize weights:
        \begin{equation}
        \Tilde{\mathbf{w}}_t^{(j)} = \frac{\mathbf{w}_t^{(j)}}{\sum_{j=1}^J \mathbf{w}_t^{(j)}}.
        \label{equ:normalise}
        \end{equation}

        \item Resample if $J_\mathrm{eff} < J/2$, where
        \begin{equation}
        J_\mathrm{eff} = \frac{1}{\sum_{j=1}^J (\Tilde{\mathbf{w}}_t^{(j)})^2}.
        \label{equ:jeff}
        \end{equation}
        Resampled weights are reset to uniform.

        \item Propose new particles:
        \begin{equation}
        \bm{\theta}_t^{(j)} \sim q_t(\cdot \mid \bm{\theta}_{t-1}^{(j)}),
        \end{equation}
        and update weights:
        \begin{equation}
        \mathbf{w}_t^{(j)} = \mathbf{w}_{t-1}^{(j)} p(\mathcal{D} \mid \bm{\theta}_t^{(j)}).
        \label{equ:chopin_wu}
        \end{equation}
    \end{enumerate}
\end{enumerate}

A detailed introduction to SMC samplers is provided in~\cite{delmoral2002sequential}, and an implementation outline appears in Algorithm~\ref{alg:base_example} in Appendix~\ref{app:smc_pseudo}.

\section{Stochastic Gradient Hamiltonian Monte Carlo} \label{sec:hmc}
This section describes how we use the Stochastic Gradient Hamiltonian Monte Carlo (SGHMC) algorithm as a proposal to help scale SMC Samplers to Bayesian Neural Networks. An overview of HMC dynamics is given is section \ref{app:hmc_dynamics}. 

\subsection{Applying SMC Samplers to Bayesian Neural Networks}
The posterior distribution over the network parameters $\bm{\theta}$ is:
\begin{equation}
\pi(\bm{\theta}) = p(\mathbf{y}_{1:N}|\bm{\theta}) q_0(\bm{\theta}) \propto p(\bm{\theta} \mid \mathbf{y}_{1:N}),
\end{equation}

where $\mathbf{y}_{1:N}$ denotes the full dataset, $q_0(\bm{\theta})$ is the prior distribution, and $p(\mathbf{y}_{1:N}|\bm{\theta})$ is the likelihood modeled by the neural network.

In standard HMC, computing the log-posterior and its gradient with respect to the entire dataset is often infeasible for large datasets and high-dimensional models. To address this, we use mini-batch stochastic gradient descent (SGD) techniques~\cite{amari1993backpropagation} to approximate full gradients. When using mini-batches of size $M$, we approximate the full-batch gradient as:
\begin{equation}
\nabla \log p(\mathbf{y}_{1:N} \mid \bm{\theta}) \approx \frac{N}{M} \mathbb{E}\left[\nabla \log p(\mathbf{y}_{\mathbf{S}_M} \mid \bm{\theta})\right],
\label{eq:smcsghmc_mb_approx}
\end{equation}

where $\mathbf{S}_M$ denotes the mini-batch indices, and $\mathbf{y}_{\mathbf{S}_M}$ is the corresponding subset of data. This approximation, given by Eq.~\eqref{eq:smcsghmc_mb_approx}, is an unbiased estimator of the true gradient. A proof is provided in Appendix~\ref{app:unbiased_grads}.  
We therefore approximate the gradient of the negative log-posterior as:
\begin{equation}
\nabla \tilde{U}(\mathbf{y}_{\mathbf{S}_M} , \bm{\theta}) = \frac{N}{M} \mathbb{E}\left[\nabla \log p(\mathbf{y}_{\mathbf{S}_M} \mid \bm{\theta})\right] - \nabla \log q_0(\bm{\theta}),
\end{equation}

where the second term accounts for the prior contribution.

\subsection{Stochastic Gradient Hamiltonian Dynamics for Mini-Batches}

\begin{algorithm}
    \caption{SGHMC Proposal for Mini-Batches}
    \label{alg:SGHMC_proposal}
    \begin{algorithmic}[1]
        \STATE \textbf{Inputs:} $\bm{\theta} = (\theta^j)_{j=1}^{J}$ (current samples), $\epsilon$ (step size), dataset $\mathcal{D} = \{(\mathbf{x}_n, \mathbf{y}_n)\}_{n=1}^N$
        \STATE \textbf{Initialize:} mini-batch indices $\mathbf{S}_M \in \mathbb{Z}^M$ 
        \FOR{each $\theta^j$ in $\bm{\theta}$}
            \STATE $\theta \gets \theta^j$
            \STATE $r \sim \mathcal{N}(0, I)$ \COMMENT{Sample initial momentum}
            \STATE Estimate initial stochastic gradient: $\nabla \tilde{U}(\mathbf{y}_{\mathbf{S}_M(1)} , \theta)$
            \STATE $r \gets r + \frac{\epsilon}{2} \nabla \tilde{U}(\mathbf{y}_{\mathbf{S}_M(1)} , \theta)$ 
            \FOR{$t = 1$ to $|\mathbf{S}_M|$}
                \STATE $\theta \gets \theta + \epsilon r$
                \IF{$t < |\mathbf{S}_M|$}
                    \STATE Estimate stochastic gradient at new $\theta$: $\nabla \tilde{U}(\mathbf{y}_{\mathbf{S}_M(t+1)} , \theta)$
                    \STATE $r \gets r + \epsilon \nabla \tilde{U}(\mathbf{y}_{\mathbf{S}_M(t+1)} , \theta)$
                \ENDIF
            \ENDFOR
            \STATE Estimate final stochastic gradient at $\theta$: $\nabla \tilde{U}(\mathbf{y}_{\mathbf{S}_M(\text{end})} , \theta)$
            \STATE $r \gets r + \frac{\epsilon}{2} \nabla \tilde{U}(\mathbf{y}_{\mathbf{S}_M(\text{end})} , \theta)$
        \ENDFOR
        \STATE \textbf{Return:} Updated sample set $\bm{\theta}$
    \end{algorithmic}
\end{algorithm}

Algorithm~\ref{alg:SGHMC_proposal} shows how each trajectory is constructed over multiple leapfrog steps, where each leapfrog step uses a different mini-batch. Thus, a single trajectory corresponds to one pass through the entire dataset, analogous to an epoch in SGD training.

Although we use mini-batches for position updates, the SMC weight update relies on full-dataset log-probability evaluations. Since weight updates only require loss evaluations (not gradients), efficient vectorized computation (e.g., via JAX~\cite{jax2018github}) significantly mitigates computational overhead.

\subsection{Entropy-Preserving SGHMC Dynamics}

In~\cite{chen2014stochastic}, SGHMC introduces noise to correct for mini-batch stochasticity:
\begin{equation}
\nabla \tilde{U}(\bm{\theta}) \approx \nabla U(\bm{\theta}) + \mathcal{N}(0, V(\bm{\theta})),
\end{equation}

where $V(\bm{\theta})$ represents gradient noise covariance. The resulting SGHMC dynamics include a friction term:
\begin{equation}
    \begin{cases}
        d\bm{\theta} = M^{-1} \mathbf{r} \, dt, \\
        d\mathbf{r} = -\nabla U(\bm{\theta}) \, dt - B(\bm{\theta}) M^{-1} \mathbf{r} \, dt + \mathcal{N}(0, 2B(\bm{\theta}) \, dt),
    \end{cases}
    \label{equ:sto_equ_friction}
\end{equation}

where $B(\bm{\theta}) = \frac{1}{2}V(\bm{\theta})$. In our variant, we omit noise injection and friction correction entirely. This leads to entropy-preserving dynamics:
\begin{equation}
    \frac{\partial}{\partial t} h(p_t(\bm{\theta}, \mathbf{r})) = - \frac{\partial}{\partial t} \int_{\bm{\theta}, \mathbf{r}} f(p_t(\bm{\theta}, \mathbf{r})) \, d\bm{\theta} \, d\mathbf{r} = 0,
\end{equation}

where $f(x) = x \ln{x}$ is a strictly convex function. A full proof is given in Appendix~\ref{app:entropy}, offering a simplified version of the derivation in~\cite{chen2014stochastic}.

\section{Scalable Proposals for Bayesian Neural Networks} \label{sec:scalable}

\subsection{Warm-up}
Warm-up is a common procedure used in the Bayesian inference literature to calibrate sampling performance. Typically, warm-up is employed to adapt hyperparameters such as the step size, mass matrix, and trajectory length~\cite{hoffman2011nouturnsampleradaptivelysetting, hoffman2021adaptive, buchholz2020adaptive}. Adapting these hyperparameters during the main sampling phase can violate the stationarity, as it may cause the sampler dynamics to lose reversibility. In the SGLD and SGHMC literature, warm-up has similarly been used to adapt other quantities, such as the noise covariance $V(\bm{\theta})$~\cite{kim2020stochasticgradientlangevindynamics} and the step size~\cite{zhang2020cyclicalstochasticgradientmcmc}.

In our setting, we do not adapt any hyperparameters during sampling. Nevertheless, we still use a warm-up phase, but for a different purpose. At initialization, the particles are unlikely to be representative of the target posterior distribution. Collecting samples from these regions would be unnecessary and potentially harmful to inference quality. Therefore, the warm-up phase in our context is used purely to allow the particles to explore the distribution and move toward regions of higher posterior density. This helps ensure that when sample collection begins, the particles have reached areas that are more informative about the posterior, even if they are not exactly located at the global modes.

\subsubsection{Targeting Tempered Posteriors}

Tempering is a common strategy in Sequential Monte Carlo (SMC) samplers, where the likelihood contribution to the posterior is introduced gradually~\cite{zhang2020cyclicalstochasticgradientmcmc, li2016learning, fortunato2019bayesianrecurrentneuralnetworks}. A standard approach is to adapt the temperature $\beta_m$ of the sampler from $0$ to $1$~\cite{del2006sequential, buchholz2020adaptive}. Often we define a sequence of tempered distributions bridging the prior and the posterior:
\begin{equation}
    p_T(\bm{\theta} \mid \mathcal{D}) \propto p(\mathcal{D} \mid \bm{\theta})^{\beta_m} p(\bm{\theta}) ,\text{for } m = 0, 1, \dots, M
\end{equation}

where \( \beta_0 = 0 < \beta_1 < \cdots < \beta_M = 1 \) is called the \textit{tempering schedule} and $M$ is the number of warm up iterations. Other approaches such as \textit{adaptive tempering} \cite{buchholz2020adaptive} have also been applied. 

However, it has been previously noted that this procedure can cause significant issues in high-dimensional settings, leading to severe particle degeneracy. In our experiments, this hypothesis was confirmed: temperature adaptation consistently resulted in an effective sample size (ESS) of $1$, implying that only a single particle meaningfully contributed to the posterior estimate. The details of these experiments can be found in Appendix~\ref{app:untempered_smc}.

To address this, the concept of \emph{cold posteriors} has been introduced in the literature~\cite{aitchison2021statisticaltheorycoldposteriors, bachmann2022temperingfixesdataaugmentation, wenzel2020goodbayesposteriordeep, kapoor2022uncertaintytemperingdataaugmentation}. Cold posteriors mitigate degeneracy by flattening the likelihood surface, allowing a greater plurality of particles to contribute to the posterior estimate. The tempered posterior $p_T(\bm{\theta} \mid \mathcal{D})$ uses a constant value of $\beta$ instead and can be expressed as:
\begin{equation}
    p_T(\bm{\theta} \mid \mathcal{D}) \propto p(\mathcal{D} \mid \bm{\theta})^{1/T} p(\bm{\theta}) ,
\end{equation}

Where $\beta = \frac{1}{T}$. The associated SMC sampler weight update is given by:
\begin{equation}
    \mathbf{w}_k^{(i)} = \mathbf{w}_{k-1}^{(i)} \left( p(\mathcal{D} \mid \bm{\theta}^{(i)}) \right)^{1/T}.
    \label{equ:tempered_wu}
\end{equation}

We set the temperature to be $T = |\mathcal{D}|$, the size of the training dataset. This corresponds to computing the mean of the log-likelihood terms rather than the sum, a common practice in SGD-based optimization to stabilize gradients:
\begin{equation}
    \log \left( p(\mathcal{D} \mid \bm{\theta}^{(i)})^{1/T} \right) = \frac{1}{|\mathcal{D}|} \sum_{d=1}^{|\mathcal{D}|} \log \left( p(\mathbf{y}_d \mid \mathbf{x}_d, \bm{\theta}^{(i)}) \right).
\end{equation}

We acknowledge that this setting means we are no longer exactly targeting the true posterior. However, in practice, it yields strong predictive performance and outperforms conventional ensemble methods, as we demonstrate in later sections. Thus, in our framework, we treat the tempered distribution as the effective target distribution for inference.

\begin{tcolorbox}[colback=gray!10, colframe=black!50, title=Remark: Cold Posterior Trick]
Tempering the posterior with $T \gg 1$ flattens the likelihood surface and alleviates particle degeneracy in high-dimensional settings.
While this means the sampler no longer exactly targets the true posterior, it enables practical and scalable inference by maintaining a diverse particle population.
Thus, the tempered posterior is treated as the effective target distribution.
\end{tcolorbox}

\subsection{Leveraging SGD Initialization and Comparison to Other Bayesian Methods}
\label{sec:sgd_transfer}

For larger models, we observed that the dynamics of the sampler converged slowly toward regions of high posterior probability. To accelerate this convergence, we leverage aspects of stochastic gradient descent (SGD) and transfer learning~\cite{hosna2022transfer}. Drawing samples is most effective when the sampler is near the high-density regions of the posterior, but reaching these regions directly via random initialization can be computationally expensive and time-consuming. Therefore, we adopt a three-stage strategy to improve sampling efficiency:

\begin{enumerate}
    \item \textbf{Pretraining with SGD:} We first train a model (or ensemble of models) using standard SGD techniques. Once training converges, the resulting weights and biases are used to initialize the warm-up phase of the sampler. Alternatively, transfer learning techniques can be employed to reuse pretrained weights as starting points.

    \item \textbf{Warm-up Sampling:} We run the sampler initialized from the pretrained model to explore the posterior distribution around the optimum found by SGD. This exploration can help improve generalization and potentially counteract overfitting introduced during the SGD phase. This mechanism is discussed further in Appendix~\ref{app:unlearn_overfit}.

    \item \textbf{Posterior Sampling:} After sufficient exploration during warm-up, we begin collecting samples from the sampler to build an approximation of the posterior distribution. These samples are aggregated at the end of training to form the posterior estimate.
\end{enumerate}

This strategy also has precedent in prior work. In \cite{zhang2020cyclicalstochasticgradientmcmc}, it was noted that when testing on larger datasets and models, noise injection was delayed until partway through the sampling process. This is significant because in~\cite{chen2014stochastic}, it was shown that SGHMC without noise and at $T=0$ reduces to standard SGD with momentum. Thus, this work was effectively using SGD dynamics to reach regions of high posterior density efficiently before beginning proper sampling.

The key difference between our approach and earlier methods is that our framework is agnostic to the choice of SGD routine used during pretraining. We simply require a method that produces effective initial weights. For example, in our CIFAR-10 and CIFAR-100 experiments, we used the AdamW optimizer \cite{loshchilov2019decoupledweightdecayregularization} with a decaying learning rate schedule. This design offloads much of the computational burden to the pretraining phase, taking advantage of extensive research into scalable frequentist optimization techniques.

Other research has similarly adopted strategies involving post-SGD sampling. Stochastic Weight Averaging Gaussian (SWAG)~\cite{maddox2019simplebaselinebayesianuncertainty} explores modes around an SGD optimum via post-training SGD runs with an increased learning rate, then fits a Gaussian approximation to the samples. Laplace Approximation (LA) methods~\cite{ritter2018scalable, daxberger2022laplacereduxeffortless} use a pretrained network to compute the Hessian (or Fisher Information Matrix) at the optimum and build a Gaussian approximation of the posterior.

In contrast, our method allows the unadjusted HMC dynamics to explore the posterior directly post-training. Instead of fitting a parametric Gaussian approximation, we form a nonparametric weighted population approximation using samples drawn during the sampling phase. Algorithm~\ref{alg:SMCSGHMC} summarizes the full SMC sampler with HMC-inspired proposals:
\begin{algorithm}[tb]
   \caption{SMC sampler for $T$ iterations and $J$ samples.}
    \begin{algorithmic}
        \STATE \textbf{Inputs:} $N$ (samples), $K$ (epochs), $B$ (warm-up epochs), $T_M, T_B$ (tempering constants), $\epsilon$ (step size), dataset $\mathcal{D} = \{(\mathbf{x}_n, \mathbf{y}_n)\}_{n=1}^N$
        \STATE \textbf{Initialize:} $\text{samples} \gets []$ \COMMENT{Initialize empty list to store samples}
        \IF{Network is pretrained}
            \STATE $\theta_0 \gets \theta_{\text{pretrained}}$ \COMMENT{Starting samples are pretrained parameters}
        \ELSE
            \STATE Sample $\{\bm{\theta}^{(j)}_0\}^J_{j=1} \sim q_0(\cdot)$
        \ENDIF
        \STATE Set initial weights $\mathbf{w}_0^{(j)}$ using \eqref{equ:init_sample}
        \FOR{$k=1$ {\bfseries to} $K$}
            \FOR{$j=1$ {\bfseries to} $J$}
                \STATE Normalise weights using \eqref{equ:normalise}
                \ENDFOR
                \STATE Calculate $J_\mathrm{eff}$ using \eqref{equ:jeff}
                \IF{$J_\mathrm{eff} < J/2$}
                \STATE Resample $[\bm{\theta}^{(1)}_k ... \bm{\theta}^{(J)}_k]$ with probability $[\Tilde{\mathbf{w}}^{(1)}_t ... \Tilde{\mathbf{w}}^{(J)}_t]$
                \STATE Reset all weights to $\frac{1}{J}$
                \ENDIF
                \FOR{$j=1$ {\bfseries to} $J$}
                \STATE $\theta^j$ = $SGHMC(\theta^{j-1}, \epsilon, \mathcal{D})$ \COMMENT{Using algorithm \ref{alg:SGHMC_proposal}}
                \STATE Update $\mathbf{w}_t^{(j)}$ using \eqref{equ:tempered_wu} with $T_M$ if $k > B$, else $T_B$
                \IF{$k > B$}
                    \STATE Store $\bm{\theta}_t^{(j)}$ in \texttt{samples}
                \ENDIF
            \ENDFOR
        \ENDFOR
    \end{algorithmic}
    \label{alg:SMCSGHMC}
\end{algorithm}

\section{Experiments}\label{sec:experiments}
This section describes the experimental setup used to evaluate our method. For the MNIST~\cite{lecun1998mnist} and FashionMNIST~\cite{xiao2017fashionmnist} image classification tasks, models were trained from scratch using the proposed SMC framework. For more complex neural networks, such as those used in the CIFAR-10 and CIFAR-100 experiments, we observed that direct training with SMC resulted in slow convergence.  Therefore, we employed the warm-up strategy outlined in Section~\ref{sec:sgd_transfer}, where models are first pretrained using standard optimization techniques before applying SMC sampling. Similarly, for the Out-of-Distribution (OOD) experiments, where larger and deeper architectures were used, we adopted the same pretraining and warm-up procedure to facilitate faster convergence and better posterior exploration. Further details regarding model architectures, hyperparameters, and training protocols are provided in Appendix~\ref{app:supplement_experi} and the weighted ensemble for SMC samples is described in \ref{app:weighted_ensemble}.

\subsection{Synthetic Multimodal Data} \label{sec:synthetic_multimodal}
\begin{figure}
    \begin{minipage}{0.98\textwidth}
        \centering
        \includegraphics[width=1.0\textwidth]{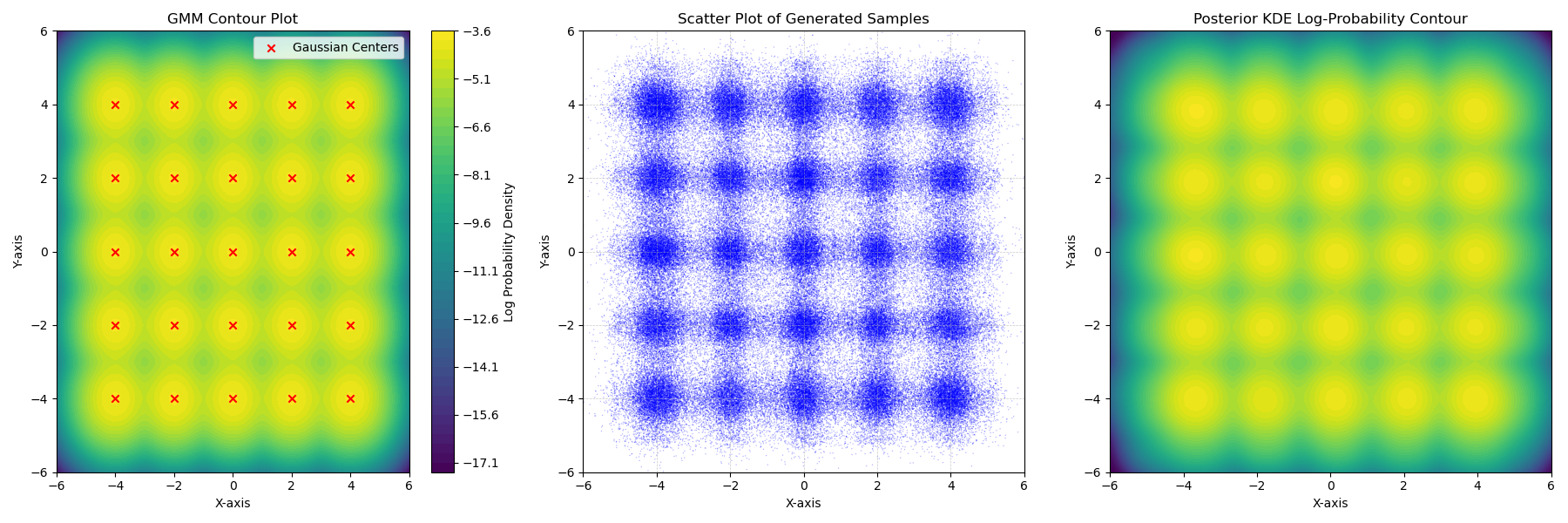} 
        \caption{Left: 25 Mode GMM Distribution which we are targeting. Center: Samples generated by the SMC sampler with a HMC proposal, a total of 200,000 samples were collected. Right: Kernel Density Estimate of the posterior distribution based off the samples generated.}
    \end{minipage}
    \label{fig:gmm_samples}
\end{figure}

To demonstrate the ability of our HMC-based proposal to approximate complex posterior distributions, we first consider sampling from a synthetic multimodal target distribution. A similar setup was considered in~\cite{zhang2020cyclicalstochasticgradientmcmc}. The target distribution is a two-dimensional mixture of 25 equally weighted Gaussian components, arranged on a grid. Each component has a diagonal covariance matrix with variance $0.3$ along both axes. The parameters of this experiment are detailed in \ref{app:experi_setup}. Figure~\ref{fig:gmm_samples} shows the contour plot of the target distribution alongside samples generated by our method. We observe that the sampler successfully explores all modes without difficulty, capturing the full structure of the posterior distribution.

\subsection{Bayesian Neural Networks for Image Classification} \label{sec:class_results_experi}
Table~\ref{tab:image_results} summarizes the results on the four image classification benchmarks described previously.  
We report the mean test loss, test accuracy, and expected calibration error (ECE)~\cite{guo2017calibrationmodernneuralnetworks}, along with standard deviations across multiple runs.

For the CIFAR-10 and CIFAR-100 datasets~\cite{krizhevsky2009learning}, we report metrics after both the initial SGD training phase and following refinement using the SMC sampling procedure. In addition, we compare against deep ensemble models~\cite{lakshminarayanan2017simplescalablepredictiveuncertainty}, where each ensemble consists of five models independently trained with different random initializations.

Across all datasets, we observe that SMC-refined models achieve strong test loss and accuracy, outperforming pure SGD training on the CIFAR benchmarks.  
While deep ensembles achieve the best overall test loss and accuracy on CIFAR-10 and CIFAR-100, their calibration—as measured by ECE—is consistently worse compared to SMC-refined models. This indicates that although deep ensembles produce competitive predictions, their predicted probabilities are less well aligned with observed outcomes, resulting in poorer calibration.

\begin{table}[ht!]
    \centering
    \caption{Image classification results for experiments.}
    \label{tab:image_results}
    \resizebox{\textwidth}{!}{
    \begin{tabular}{llccccc}
        \toprule
        \textbf{Dataset} & \textbf{Metric} & \textbf{SMC} & \textbf{SGD} & \textbf{Deep Ensemble} & \textbf{SWAG Diag } & \textbf{SWAG}  \\
        \midrule
        MNIST & Accuracy  & 99.22 (0.04)\% & -- & -- & -- & -- \\
                  & Loss & 0.0216 (0.0002) & -- & -- & -- & -- \\
                  & ECE & 0.0083 (0.0001) & -- & -- & -- & -- \\ 
        \midrule
        FashionMNIST & Accuracy  & 92.23 (0.06)\% & -- & -- & -- & -- \\
                  & Loss & 0.2426 (0.0017) & -- & -- & -- & -- \\
                  & ECE & 0.0614 (0.0008) & -- & --  & -- & -- \\
        \midrule
        CIFAR10 & Accuracy  & 91.18 (0.32) & 90.71 (0.30)\% & \textbf{93.69 \% (0.13)}  & 90.73 (0.36) \% & 90.73 (0.30) \% \\
                  & Loss & 0.3086 (0.0072) & 0.4285 (0.0082) & \textbf{0.2537 (0.73)} & 0.4216 (0.0081) & 0.4256 (0.0081) \\
                  & ECE & \textbf{0.0499 (0.0017)} & 0.0657 (0.0019) & 0.0544 (0.0011)  & 0.0649 (0.0016) & 0.0653 (0.0015) \\
        \midrule
        CIFAR100 & Accuracy  & 71.29 (0.27) \% & 71.58 (0.39) \% & \textbf{76.34\% (0.22)}  & 71.52 (0.23) \% & 71.65 (0.35) \% \\
                  & Loss & 1.2124 (0.0163) & 1.7668 (0.0248) & \textbf{1.1692 (0.0091)} & 1.7470 (0.0428) & 1.7245 (0.0504) \\
                  & ECE & \textbf{0.1001 (0.0046)} & 0.1953 (0.0018) & 0.1411 (0.0014) & 0.1935 (0.0038) & 0.1920 (0.0043) \\
        \bottomrule
    \end{tabular}
    }
\end{table}

\subsection{Bayesian Neural Networks for Out-of-Distribution Detection} \label{sec:ood_results_experi}
To further investigate the practical implications of improved calibration, we next evaluate performance on a standard out-of-distribution (OOD) detection task, where reliable uncertainty quantification is critical. In this setting, uncertainty in the model's predictive logits is used to distinguish between in-distribution and out-of-distribution samples~\cite{lakshminarayanan2017simplescalablepredictiveuncertainty}.

We conducted experiments using several popular OOD datasets. Specifically, we trained a VGG16 neural network on the CIFAR-10 dataset and evaluated its performance against OOD samples drawn from three datasets: CIFAR-100, SVHN~\cite{goodfellow2014multidigitnumberrecognitionstreet}, and the subsets "church" and "bridge" of the LSUN dataset~\cite{yu2016lsunconstructionlargescaleimage}. For the LSUN subsets, we randomly selected 5,000 images for validation and 10,000 images for testing, matching the sample sizes used for SVHN and CIFAR-100.

The validation datasets were used to determine the decision threshold based on the FPR95 metric (False Positive Rate at 95\% True Positive Rate).  
We employ energy-based scoring~\cite{liu2021energybasedoutofdistributiondetection} to define separation between inlier and outlier classes by evaluating the model's energy function over the logits. A detailed overview of the energy scoring method is provided in Appendix~\ref{app:energy_scores}.

Table~\ref{tab:ood} presents the quantitative results. From the table we observe that the SMC-refined networks consistently achieve superior AUROC scores compared to both SGD-trained and deep ensemble models. This result demonstrates that the improved calibration observed in earlier experiments translates into tangible benefits for uncertainty quantification, leading to more reliable OOD detection performance. In Appendix \ref{app:auroc_curves}, Figures~\ref{fig:auroc_svhn}--\ref{fig:auroc_lsun_bridge} show the corresponding AUROC curves, with shaded regions representing standard deviations across runs. 

\begin{table}[ht!]
    \centering
    \caption{Performance metrics for the methods across OOD tasks.}
    \label{tab:ood}
    \resizebox{\textwidth}{!}{
    \begin{tabular}{llcccccc}
        \toprule
        \textbf{Dataset} & \textbf{Method} & \textbf{Accuracy} & \textbf{Precision} & \textbf{Recall} & \textbf{F1} & \textbf{Specificity} & \textbf{AUROC} \\
        \midrule
        \multirow{5}{*}{SVHN} 
        & SMC       & \textbf{92.41 ± 0.57} & \textbf{92.56 ± 0.44} & 92.24 ± 0.95 & \textbf{92.40 ± 0.60} & \textbf{92.59 ± 0.45} & \textbf{97.27 ± 0.40} \\
        & SGD       & 90.19 ± 1.00 & 90.02 ± 0.37 & 90.40 ± 2.05 & 90.20 ± 1.10 & 89.97 ± 0.39 & 95.25 ± 1.41 \\
        & Ensemble  & 91.98 ± 1.32 & 89.86 ± 0.38 & \textbf{94.62 ± 2.54} & 92.17 ± 1.39 & 89.33 ± 0.26 & 96.25 ± 1.14 \\
        & SWAG Diag & 91.68 ± 1.17 & 90.19 ± 0.44 & 93.53 ± 2.61 & 91.81 ± 1.26 & 89.82 ± 0.61 & 96.32 ± 1.12 \\
        & SWAG      & 90.37 ± 0.90 & 90.06 ± 0.36 & 90.76 ± 1.77 & 90.40 ± 0.98 & 89.98 ± 0.35 & 95.57 ± 1.36 \\
        \midrule
        \multirow{5}{*}{CIFAR100}
        & SMC       & 75.83 ± 0.99 & \textbf{88.92 ± 0.58} & 59.04 ± 2.69 & 70.92 ± 1.77 & \textbf{92.63 ± 0.74} & \textbf{90.39 ± 0.12} \\
        & SGD       & 75.07 ± 0.37 & 85.66 ± 0.21 & 60.21 ± 1.01 & 70.71 ± 0.65 & 89.92 ± 0.31 & 84.32 ± 0.33 \\
        & Ensemble  & \textbf{79.62 ± 0.15} & 86.83 ± 0.25 & \textbf{69.83 ± 0.45} & \textbf{77.41 ± 0.22} & 89.40 ± 0.28 & 88.83 ± 0.17 \\
        & SWAG Diag & 76.03 ± 0.39 & 85.72 ± 0.26 & 62.48 ± 0.91 & 72.27 ± 0.61 & 89.59 ± 0.27 & 87.34 ± 0.44 \\
        & SWAG      & 75.20 ± 0.55 & 85.82 ± 0.22 & 60.38 ± 1.46 & 70.88 ± 0.97 & 90.02 ± 0.40 & 85.99 ± 0.21 \\
        \midrule
        \multirow{5}{*}{LSUN Church}
        & SMC       & 79.46 ± 2.47 & \textbf{90.11 ± 0.90} & 66.14 ± 4.93 & 76.20 ± 3.64 & \textbf{92.78 ± 0.48} & \textbf{92.06 ± 1.05} \\
        & SGD       & 75.87 ± 1.18 & 85.67 ± 0.65 & 62.13 ± 2.47 & 72.00 ± 1.79 & 89.62 ± 0.50 & 84.75 ± 2.08 \\
        & Ensemble  & \textbf{79.07 ± 1.39} & 86.73 ± 0.56 & \textbf{68.61 ± 2.72} & \textbf{76.59 ± 1.89} & 89.52 ± 0.25 & 88.95 ± 1.84 \\
        & SWAG Diag & 74.89 ± 1.78 & 85.44 ± 1.02 & 59.97 ± 3.39 & 70.44 ± 2.69 & 89.81 ± 0.37 & 85.69 ± 2.74 \\
        & SWAG      & 75.36 ± 1.15 & 85.85 ± 0.68 & 60.71 ± 2.42 & 71.10 ± 1.77 & 90.00 ± 0.53 & 85.56 ± 2.06 \\
        \midrule
        \multirow{5}{*}{LSUN Bridge}
        & SMC       & 77.88 ± 2.20 & \textbf{89.68 ± 0.87} & 62.98 ± 4.42 & 73.92 ± 3.35 & \textbf{92.78 ± 0.48} & \textbf{91.65 ± 0.97} \\
        & SGD       & 76.42 ± 0.72 & 85.89 ± 0.58 & 63.23 ± 1.48 & 72.83 ± 1.06 & 89.62 ± 0.50 & 85.77 ± 1.95 \\
        & Ensemble  & \textbf{80.00 ± 0.90} & 87.05 ± 0.41 & \textbf{70.48 ± 1.74} & \textbf{77.89 ± 1.19} & 89.52 ± 0.25 & 90.22 ± 1.23 \\
        & SWAG Diag & 75.65 ± 1.14 & 85.77 ± 0.71 & 61.49 ± 2.14 & 71.61 ± 1.66 & 89.81 ± 0.37 & 87.06 ± 2.16 \\
        & SWAG      & 75.99 ± 0.81 & 86.11 ± 0.57 & 61.98 ± 1.81 & 72.07 ± 1.24 & 90.00 ± 0.53 & 86.85 ± 1.89 \\
        \bottomrule
    \end{tabular}}
\end{table}

\subsection{Transfer Learning with the STL Dataset} \label{sec:transfer_results_experi}
We also evaluated all methods on a standard transfer learning task. Specifically, we used the STL-10 dataset~\cite{coates2011analysis}, an image classification dataset containing 10 classes derived from ImageNet, similar to CIFAR-10. The STL-10 and CIFAR-10 datasets share nine out of ten classes; the primary difference is that CIFAR-10 includes a ``frog'' class, while STL-10 includes a ``monkey'' class. It is also important to note that although the class names largely overlap, the actual image samples differ between the two datasets.

For this evaluation, we reused the VGG16 networks previously trained on CIFAR-10 during the OOD experiments for each method (SGD, deep ensemble, and SMC-refined models). The STL-10 dataset was introduced at test time to assess how well each method generalizes to a new, but semantically similar, distribution. The results are reported in Table~\ref{tab:transfer_results}.

\begin{table}[ht!]
    \centering
    \caption{Image classification results for the transfer learning task.}
    \label{tab:transfer_results}
    \resizebox{\textwidth}{!}{
    \begin{tabular}{lccccc}
        \toprule
        \textbf{Metric} & \textbf{SMC} & \textbf{SGD} & \textbf{Deep Ensemble} & \textbf{SWAG Diag} & \textbf{SWAG}  \\
        \midrule
        Accuracy  & 71.13 (0.63)\% & 71.83 (0.36)\% & \textbf{73.74 (0.35)\%} & 71.78 (0.34) \% & 71.85 (0.26) \% \\
        Loss & \textbf{0.9136 (0.0267)} & 1.7851 (0.0245) & 1.4443 (0.0151) & 1.7347 (0.0206) & 1.7275 (0.0181) \\
        ECE & \textbf{0.0745 (0.0101)} & 0.2388 (0.0039) & 0.2077 (0.0031) & 0.2369 (0.0033) & 0.2354 (0.0024) \\ 
        \bottomrule
    \end{tabular}
    }
\end{table}

\section{Discussion} \label{sec:results}
In this work, we have demonstrated that a Stochastic Gradient Hamiltonian Monte Carlo (SGHMC) proposal can be effectively incorporated within a Sequential Monte Carlo (SMC) sampler, enabling scalable Bayesian inference for complex neural network posteriors. By applying our method to pretrained neural networks, we showed that it is possible to mitigate overfitting effects, leading to improved generalization performance.

Specifically, we trained neural networks using standard SGD, then applied the SMC sampling procedure as a post-training refinement phase. This conversion from deterministic neural networks (NNs) to Bayesian neural networks (BNNs) consistently improved test loss, accuracy, and expected calibration error (ECE) compared to SGD alone. Moreover, we demonstrated that improved calibration translates to tangible benefits on uncertainty quantification tasks and transfer learning benchmarks, outperforming established methods such as Deep Ensembles, SWAG, and Diagonal SWAG. We also provided a complete training pipeline for converting pretrained NNs into calibrated BNNs in a principled manner.

\subsection{Further Work} \label{sec:further_work}
While the proposed approach achieves strong empirical results, it introduces additional computational cost compared to standard SGD training. Furthermore, we observed that if the sampler is run for too long, the validation loss can begin to degrade, suggesting that particles may drift into non-optimal regions of the posterior. In our experiments, we addressed this manually by monitoring validation loss and terminating sampling early. Developing automated stopping criteria based on validation metrics or entropy measures remains an interesting direction for future research.

Finally, extending this framework to larger-scale settings—such as training on the full ImageNet dataset or applying it to modern transformer architectures—would further validate the scalability and robustness of the method.


\begin{thebibliography}{10}

\bibitem{aitchison2021statisticaltheorycoldposteriors}
Laurence Aitchison.
\newblock A statistical theory of cold posteriors in deep neural networks, 2021.

\bibitem{amari1993backpropagation}
Shun-ichi Amari.
\newblock Backpropagation and stochastic gradient descent method.
\newblock {\em Neurocomputing}, 5(4-5):185--196, 1993.

\bibitem{bachmann2022temperingfixesdataaugmentation}
Gregor Bachmann, Lorenzo Noci, and Thomas Hofmann.
\newblock How tempering fixes data augmentation in bayesian neural networks, 2022.

\bibitem{bai2021understandingimprovingearlystopping}
Yingbin Bai, Erkun Yang, Bo~Han, Yanhua Yang, Jiatong Li, Yinian Mao, Gang Niu, and Tongliang Liu.
\newblock Understanding and improving early stopping for learning with noisy labels, 2021.

\bibitem{jax2018github}
James Bradbury, Roy Frostig, Peter Hawkins, Matthew~James Johnson, Chris Leary, Dougal Maclaurin, George Necula, Adam Paszke, Jake Vander{P}las, Skye Wanderman-{M}ilne, and Qiao Zhang.
\newblock {JAX}: composable transformations of {P}ython+{N}um{P}y programs, 2018.

\bibitem{Brooks_2011}
Steve Brooks, Andrew Gelman, Galin Jones, and Xiao-Li Meng.
\newblock {\em Handbook of Markov Chain Monte Carlo}.
\newblock Chapman and Hall/CRC, May 2011.

\bibitem{buchholz2020adaptive}
Alexander Buchholz, Nicolas Chopin, and Pierre~E. Jacob.
\newblock Adaptive tuning of hamiltonian monte carlo within sequential monte carlo, 2020.

\bibitem{chen2014stochastic}
Tianqi Chen, Emily Fox, and Carlos Guestrin.
\newblock Stochastic gradient hamiltonian monte carlo.
\newblock In {\em International conference on machine learning}, pages 1683--1691. PMLR, 2014.

\bibitem{chen2014stochasticgradienthamiltonianmonte}
Tianqi Chen, Emily~B. Fox, and Carlos Guestrin.
\newblock Stochastic gradient hamiltonian monte carlo, 2014.

\bibitem{chopin2020introduction}
Nicolas Chopin, Omiros Papaspiliopoulos, et~al.
\newblock {\em An introduction to sequential Monte Carlo}, volume~4.
\newblock Springer, 2020.

\bibitem{coates2011analysis}
Adam Coates, Andrew Ng, and Honglak Lee.
\newblock An analysis of single-layer networks in unsupervised feature learning.
\newblock In {\em Proceedings of the fourteenth international conference on artificial intelligence and statistics}, pages 215--223. JMLR Workshop and Conference Proceedings, 2011.

\bibitem{daxberger2022laplacereduxeffortless}
Erik Daxberger, Agustinus Kristiadi, Alexander Immer, Runa Eschenhagen, Matthias Bauer, and Philipp Hennig.
\newblock Laplace redux -- effortless bayesian deep learning, 2022.

\bibitem{DelMoral2006}
Pierre Del~Moral, Arnaud Doucet, and Ajay Jasra.
\newblock Sequential monte carlo samplers.
\newblock {\em Journal of the Royal Statistical Society: Series B (Statistical Methodology)}, 68(3):411--436, 2006.

\bibitem{del2006sequential}
Pierre Del~Moral, Arnaud Doucet, and Ajay Jasra.
\newblock Sequential monte carlo samplers.
\newblock {\em Journal of the Royal Statistical Society Series B: Statistical Methodology}, 68(3):411--436, 2006.

\bibitem{douc2005comparison}
Randal Douc and Olivier Capp{\'e}.
\newblock Comparison of resampling schemes for particle filtering.
\newblock In {\em ISPA 2005. Proceedings of the 4th International Symposium on Image and Signal Processing and Analysis, 2005.}, pages 64--69. Ieee, 2005.

\bibitem{doucet2001introduction}
Arnaud Doucet, Nando de~Freitas, and Neil~J. Gordon.
\newblock An introduction to sequential monte carlo methods.
\newblock In Arnaud Doucet, Nando de~Freitas, and Neil~J. Gordon, editors, {\em Sequential Monte Carlo Methods in Practice}, pages 3--14. Springer, New York, NY, 2001.

\bibitem{fortunato2019bayesianrecurrentneuralnetworks}
Meire Fortunato, Charles Blundell, and Oriol Vinyals.
\newblock Bayesian recurrent neural networks, 2019.

\bibitem{goodfellow2014multidigitnumberrecognitionstreet}
Ian~J. Goodfellow, Yaroslav Bulatov, Julian Ibarz, Sacha Arnoud, and Vinay Shet.
\newblock Multi-digit number recognition from street view imagery using deep convolutional neural networks, 2014.

\bibitem{guo2017calibrationmodernneuralnetworks}
Chuan Guo, Geoff Pleiss, Yu~Sun, and Kilian~Q. Weinberger.
\newblock On calibration of modern neural networks, 2017.

\bibitem{hoffman2021adaptive}
Matthew Hoffman, Alexey Radul, and Pavel Sountsov.
\newblock An adaptive-mcmc scheme for setting trajectory lengths in hamiltonian monte carlo.
\newblock In {\em International Conference on Artificial Intelligence and Statistics}, pages 3907--3915. PMLR, 2021.

\bibitem{hoffman2011nouturnsampleradaptivelysetting}
Matthew~D. Hoffman and Andrew Gelman.
\newblock The no-u-turn sampler: Adaptively setting path lengths in hamiltonian monte carlo, 2011.

\bibitem{hosna2022transfer}
Asmaul Hosna, Ethel Merry, Jigmey Gyalmo, Zulfikar Alom, Zeyar Aung, and Mohammad~Abdul Azim.
\newblock Transfer learning: a friendly introduction.
\newblock {\em Journal of Big Data}, 9(1):102, 2022.

\bibitem{izmailov2021bayesianneuralnetworkposteriors}
Pavel Izmailov, Sharad Vikram, Matthew~D. Hoffman, and Andrew~Gordon Wilson.
\newblock What are bayesian neural network posteriors really like?, 2021.

\bibitem{johansen2008particle}
Adam~M Johansen, Arnaud Doucet, and Manuel Davy.
\newblock Particle methods for maximum likelihood estimation in latent variable models.
\newblock {\em Statistics and Computing}, 18:47--57, 2008.

\bibitem{kapoor2022uncertaintytemperingdataaugmentation}
Sanyam Kapoor, Wesley~J. Maddox, Pavel Izmailov, and Andrew~Gordon Wilson.
\newblock On uncertainty, tempering, and data augmentation in bayesian classification, 2022.

\bibitem{kim2020stochasticgradientlangevindynamics}
Sehwan Kim, Qifan Song, and Faming Liang.
\newblock Stochastic gradient langevin dynamics algorithms with adaptive drifts, 2020.

\bibitem{kloek1978bayesian}
Teun Kloek and Herman~K Van~Dijk.
\newblock Bayesian estimates of equation system parameters: an application of integration by monte carlo.
\newblock {\em Econometrica: Journal of the Econometric Society}, pages 1--19, 1978.

\bibitem{krizhevsky2009learning}
Alex Krizhevsky.
\newblock Learning multiple layers of features from tiny images.
\newblock Technical report, University of Toronto, 2009.

\bibitem{lakshminarayanan2017simplescalablepredictiveuncertainty}
Balaji Lakshminarayanan, Alexander Pritzel, and Charles Blundell.
\newblock Simple and scalable predictive uncertainty estimation using deep ensembles, 2017.

\bibitem{lecun1998mnist}
Yann LeCun, L{\'e}on Bottou, Yoshua Bengio, and Patrick Haffner.
\newblock Gradient-based learning applied to document recognition.
\newblock {\em Proceedings of the IEEE}, 86(11):2278--2324, 1998.

\bibitem{li2016learning}
Chunyuan Li, Andrew Stevens, Changyou Chen, Yunchen Pu, Zhe Gan, and Lawrence Carin.
\newblock Learning weight uncertainty with stochastic gradient mcmc for shape classification.
\newblock In {\em Proceedings of the IEEE Conference on Computer Vision and Pattern Recognition}, pages 5666--5675, 2016.

\bibitem{liu2021energybasedoutofdistributiondetection}
Weitang Liu, Xiaoyun Wang, John~D. Owens, and Yixuan Li.
\newblock Energy-based out-of-distribution detection, 2021.

\bibitem{loshchilov2019decoupledweightdecayregularization}
Ilya Loshchilov and Frank Hutter.
\newblock Decoupled weight decay regularization, 2019.

\bibitem{maceachern1999sequential}
Steven~N MacEachern, Merlise Clyde, and Jun~S Liu.
\newblock Sequential importance sampling for nonparametric bayes models: The next generation.
\newblock {\em Canadian Journal of Statistics}, 27(2):251--267, 1999.

\bibitem{maddox2019simplebaselinebayesianuncertainty}
Wesley Maddox, Timur Garipov, Pavel Izmailov, Dmitry Vetrov, and Andrew~Gordon Wilson.
\newblock A simple baseline for bayesian uncertainty in deep learning, 2019.

\bibitem{13ab5b5e-0237-33fb-a7a8-6f6e4e0d4e0f}
Nicholas Metropolis and S.~Ulam.
\newblock The monte carlo method.
\newblock {\em Journal of the American Statistical Association}, 44(247):335--341, 1949.

\bibitem{delmoral2002sequential}
Pierre~Del Moral and Arnaud Doucet.
\newblock Sequential monte carlo samplers, 2002.

\bibitem{neal2011mcmc}
Radford~M. Neal.
\newblock {\em MCMC using Hamiltonian dynamics}, pages 113--162.
\newblock Chapman and Hall/CRC, 2011.

\bibitem{ritter2018scalable}
Hippolyt Ritter, Aleksandar Botev, and David Barber.
\newblock A scalable laplace approximation for neural networks.
\newblock In {\em 6th international conference on learning representations, ICLR 2018-conference track proceedings}, volume~6. International Conference on Representation Learning, 2018.

\bibitem{ruder2017overviewgradientdescentoptimization}
Sebastian Ruder.
\newblock An overview of gradient descent optimization algorithms, 2017.

\bibitem{welling2011bayesian}
Max Welling and Yee~W Teh.
\newblock Bayesian learning via stochastic gradient langevin dynamics.
\newblock In {\em Proceedings of the 28th international conference on machine learning (ICML-11)}, pages 681--688. Citeseer, 2011.

\bibitem{wenzel2020goodbayesposteriordeep}
Florian Wenzel, Kevin Roth, Bastiaan~S. Veeling, Jakub Świątkowski, Linh Tran, Stephan Mandt, Jasper Snoek, Tim Salimans, Rodolphe Jenatton, and Sebastian Nowozin.
\newblock How good is the bayes posterior in deep neural networks really?, 2020.

\bibitem{xiao2017fashionmnist}
Han Xiao, Kashif Rasul, and Roland Vollgraf.
\newblock Fashion-mnist: a novel image dataset for benchmarking machine learning algorithms, 2017.

\bibitem{yu2016lsunconstructionlargescaleimage}
Fisher Yu, Ari Seff, Yinda Zhang, Shuran Song, Thomas Funkhouser, and Jianxiong Xiao.
\newblock Lsun: Construction of a large-scale image dataset using deep learning with humans in the loop, 2016.

\bibitem{zhang2020cyclicalstochasticgradientmcmc}
Ruqi Zhang, Chunyuan Li, Jianyi Zhang, Changyou Chen, and Andrew~Gordon Wilson.
\newblock Cyclical stochastic gradient mcmc for bayesian deep learning, 2020.

\end{thebibliography}

\clearpage
\appendix

\section{Importance Sampling} 

\subsection{Monte Carlo Approximation} \label{app:mc_tutorial}

Suppose we have a target probability distribution $\pi(\bm{\theta})$, and we want to evaluate the expectation of a function $f$ with respect to it:

\begin{equation} 
    \mathbb{E}_{\pi}\left[ f(\bm{\theta}) \right] = \int f(\bm{\theta}) \pi(\bm{\theta}) d\bm{\theta}
\end{equation}

In practice, this expectation is often approximated using the Monte Carlo (MC) method \cite{13ab5b5e-0237-33fb-a7a8-6f6e4e0d4e0f} by drawing $N$ samples from the target distribution:

\begin{equation} 
    \{\bm{\theta}^{(i)}\}_{i=1}^N \sim \pi(\bm{\theta}), 
\end{equation}

\begin{equation} 
    \mathbb{E}_{\pi}\left[ f(\bm{\theta}) \right] \approx \frac{1}{N} \sum_{i=1}^N f(\bm{\theta}^{(i)}). 
\end{equation}

This estimate converges to the true expectation as $N \to \infty$, with the variance decreasing proportionally to $1/N$.

\subsection{Importance Sampling} \label{app:is_tutorial}

When direct sampling from $\pi(\bm{\theta})$ is infeasible, we can instead use Importance Sampling (IS) \cite{kloek1978bayesian}. This technique introduces a proposal distribution $q(\bm{\theta})$ from which it is easy to generate samples, and reweights them to approximate expectations under $\pi$:

\begin{equation} 
    \mathbb{E}_{\pi}\left[ f(\bm{\theta}) \right] = \int f(\bm{\theta}) \pi(\bm{\theta})  d\bm{\theta} = \int f(\bm{\theta}) \frac{\pi(\bm{\theta})}{q(\bm{\theta})} q(\bm{\theta})  d\bm{\theta} = \mathbb{E}_{q} \left[ f(\bm{\theta}) \frac{\pi(\bm{\theta})}{q(\bm{\theta})} \right]. 
\end{equation}

The ratio $\frac{\pi(\bm{\theta})}{q(\bm{\theta})}$ is known as the importance weight, which we denote as $w(\bm{\theta})$:

\begin{equation} 
    w(\bm{\theta}) = \frac{\pi(\bm{\theta})}{q(\bm{\theta})}. 
\end{equation}

We can now approximate the expectation using samples drawn from $q(\bm{\theta})$:

\begin{equation} 
    \{\bm{\theta}^{(i)}\}_{i=1}^N \sim q(\bm{\theta}), 
\end{equation}

\begin{equation} 
    \mathbb{E}_{\pi}\left[ f(\bm{\theta}) \right] \approx \frac{1}{N} \sum_{i=1}^N f(\bm{\theta}^{(i)}) w(\bm{\theta}^{(i)}). 
\end{equation}

This yields a weighted Monte Carlo estimate, often written as:

\begin{equation} \mu_q \approx \frac{1}{N} \sum_{i=1}^N f(\bm{\theta}^{(i)}) w(\bm{\theta}^{(i)}), \end{equation}

where $\mu_q$ is an unbiased estimator of $\mu_{\pi}$, the true expectation under the target distribution. As the number of samples increases, this estimate becomes more accurate and its variance decreases. A detailed discussion of the theoretical properties of importance sampling, including proofs of unbiasedness and convergence, can be found in Appendix \ref{app:is_proofs}.

\subsection{Importance Sampling Convergence Proofs} \label{app:is_proofs}

\paragraph{Proof: $\mu_{\pi} = \mathbb{E}_q\left[\mu_q \right]$, i.e. $\mu_q$ is an unbiased estimator of $\mu_{\pi}$} 

As a note, a lot of this has been gone over in the preceding equations but I have provided the proof here for completeness sake. 

\begin{align}
    \mathbb{E}_q\left[\mu_q \right] &= \int f(\bm{\theta}) w(\bm{\theta}) q(\bm{\theta}) d\bm{\theta} \\
    & = \int f(\bm{\theta})\frac{\pi(\bm{\theta})}{q(\bm{\theta})} q(\bm{\theta}) d\bm{\theta} \\
    & = \int f(\bm{\theta}) \pi(\bm{\theta}) d\bm{\theta} \\
    & = \mathbb{E}_{\pi}\left[ f(\bm{\theta})\right] = \mu_{\pi}. 
\end{align}

\paragraph{Proof: $\mathrm{Var}_q\left[ \mu_q\right] = \frac{\sigma_q^2}{N}$}

\begin{equation}
    \mathrm{Var}_q\left[ \mu_q\right] = \mathrm{Var}_{q}\left[\frac{1}{N} \sum_{i=1}^N f(\bm{\theta}^i) w(\bm{\theta}^i)\right]\\
\end{equation}

As samples are independent and identically distributed (i.i.d), the variance of the sum is the sum of the variance. And using $\mathrm{Var}\left[aX\right] = a^2\mathrm{Var}\left[X\right]$

\begin{align}
    \mathrm{Var}_{q}\left[\frac{1}{N} \sum_{i=1}^N f(\bm{\theta}^i) w(\bm{\theta}^i)\right] &= \frac{1}{N^2} \sum_{i=1}^N \mathrm{Var}_{q}\left[ f(\bm{\theta}) w(\bm{\theta})\right] \\
    & = \frac{N}{N^2} \mathrm{Var}_{q}\left[ f(\bm{\theta}) w(\bm{\theta})\right] \\
    & = \frac{1}{N} \mathrm{Var}_{q}\left[ f(\bm{\theta}) w(\bm{\theta})\right]
\end{align}

Now we can use the variance identity $\mathrm{Var}_{q}\left[ f(\bm{\theta}) w(\bm{\theta})\right] = \mathbb{E}_{q}\left[ (f(\bm{\theta}) w(\bm{\theta}))^2\right] - \left(\mathbb{E}_{q}\left[ f(\bm{\theta}) w(\bm{\theta})\right]\right)^2$

\begin{align}
    \mathrm{Var}_{q}\left[ f(\bm{\theta}) w(\bm{\theta})\right] &= \int f(\bm{\theta})^2 w(\bm{\theta})^2 q(\bm{\theta}) d\bm{\theta} - \left(\int f(\bm{\theta}) w(\bm{\theta}) q(\bm{\theta}) d\bm{\theta}\right)^2 \\
    &= \int f(\bm{\theta})^2 w(\bm{\theta})^2 q(\bm{\theta}) d\bm{\theta} - \mu_q^2 \\
    &= \int f(\bm{\theta})^2 \frac{\pi(\bm{\theta})^2}{q(\bm{\theta})} d\bm{\theta} - \mu_q^2 \\
\end{align}

Therefore we have 
\begin{equation}
    \mathrm{Var}_q\left[ \mu_q\right] = \frac{1}{N}\left[\int f(\bm{\theta})^2 \frac{\pi(\bm{\theta})^2}{q(\bm{\theta})} d\bm{\theta} - \mu_q^2 \right]
\end{equation}

Where 
\begin{equation}
    \sigma_q^2 = \int f(\bm{\theta})^2 \frac{\pi(\bm{\theta})^2}{q(\bm{\theta})} d\bm{\theta} - \mu_q^2
\end{equation}

Finally, this gives 
\begin{equation}
    \mathrm{Var}_q\left[ \mu_q\right] = \frac{1}{N}\sigma_q^2
\end{equation}

This result is important, as it shows that as $N \xrightarrow[]{} \infty$, the variance of our estimate goes to zero. Therefore, the more samples we have the closer we are to the true answer. 

\newpage

\section{SMC Sampler Pseudocode} \label{app:smc_pseudo}

For the experiments in this paper, we induce resampling every iteration as is commonly done in other algorithms such as the Bootstrap Filter \cite{doucet2001introduction}. 

\begin{algorithm}[H]
   \caption{SMC sampler for $T$ iterations and $J$ particles.}
   \label{alg:base_example}
\begin{algorithmic}
    \STATE Sample $\{\bm{\theta}_0^{(j)}\}_{j=1}^J \sim q_0(\cdot)$
    \STATE Set initial weights $\mathbf{w}_0^{(j)}$ using Eq.~\eqref{equ:init_sample}
    \FOR{$t=1$ {\bfseries to} $T$}
        \STATE Normalize weights using Eq.~\eqref{equ:normalise}
        \STATE Compute $J_\mathrm{eff}$ using Eq.~\eqref{equ:jeff}
        \IF{$J_\mathrm{eff} < J/2$}
            \STATE Resample $\{\bm{\theta}_t^{(j)}\}$ proportionally to $\{\Tilde{\mathbf{w}}_t^{(j)}\}$
            \STATE Reset all weights to $\mathbf{w}_t^{(j)} = 1/J$
        \ENDIF
        \FOR{$j=1$ {\bfseries to} $J$}
            \STATE Propose $\bm{\theta}_t^{(j)} \sim q_t(\cdot \mid \bm{\theta}_{t-1}^{(j)})$
            \STATE Update weights using Eq.~\eqref{equ:chopin_wu}
        \ENDFOR
    \ENDFOR
\end{algorithmic}
\end{algorithm}

\newpage

\section{HMC Dynamics}

\subsection{Hamiltonian Monte Carlo Dynamics} \label{app:hmc_dynamics}

The dynamics of Hamiltonian Monte Carlo (HMC)~\cite{Brooks_2011, neal2011mcmc} are governed by the following system of differential equations:
\begin{equation} 
    \begin{cases} 
        d\bm{\theta} = M^{-1} \mathbf{r} \, dt, \\ 
        d\mathbf{r} = -\nabla U(\bm{\theta}) \, dt
    \end{cases}
    \label{equ:sto_equ}
\end{equation}

where $\bm{\theta}$ represents the position, $\mathbf{r}$ the momentum, $M^{-1}$ the inverse mass matrix, $dt$ the time step, and $U(\bm{\theta})$ the potential energy (typically the negative log-posterior). These dynamics can be compactly expressed as:
\begin{equation}
    d
    \begin{bmatrix} 
        \bm{\theta} \\ 
        \mathbf{r} 
    \end{bmatrix}
    = -G \nabla H(\bm{\theta}, \mathbf{r}) \, dt,
    \label{equ:sto_sys}
\end{equation}

where the antisymmetric matrix $G$ is given by:
\begin{equation}
    G = 
    \begin{bmatrix}
        0 & -I \\ 
        I & 0
    \end{bmatrix},
\end{equation}

and the Hamiltonian gradient is:
\begin{equation}
    \nabla H(\bm{\theta}, \mathbf{r}) = 
    \begin{bmatrix}
        \nabla U(\bm{\theta}) \\ 
        M^{-1} \mathbf{r}
    \end{bmatrix}.
\end{equation}

The Hamiltonian itself is:
\begin{equation}
    H(\bm{\theta}, \mathbf{r}) = U(\bm{\theta}) + \frac{1}{2} \mathbf{r}^T M^{-1} \mathbf{r}.
\end{equation}

Expanding this system recovers the original dynamics shown in Eq.~\eqref{equ:sto_equ}.  
A key property of Hamiltonian dynamics is time-reversibility: the dynamics remain unchanged if the momentum is negated:
\begin{equation}
    d
    \begin{bmatrix}
        \bm{\theta} \\ 
        \mathbf{r}
    \end{bmatrix}
    = d
    \begin{bmatrix}
        \bm{\theta} \\ 
        -\mathbf{r}
    \end{bmatrix}.
\end{equation}

A formal derivation of this property is provided in Appendix~\ref{app:time_rev_sghmc}.

\begin{tcolorbox}[colback=gray!10, colframe=black!50, title=Remark: HMC for SMC Samplers]
Unlike traditional MCMC settings where a Metropolis-Hastings acceptance step ensures detailed balance, SMC samplers can use unadjusted Hamiltonian dynamics combined with importance weighting and resampling to correct for bias. This enables efficient proposals without rejection steps.
\end{tcolorbox}

\subsection{Time Reversibility of HMC Dynamics} \label{app:time_rev_sghmc}
The original equation is given by \ref{equ:sto_sys}, but we have restated it here for convenience: 

\begin{equation}
    d \begin{bmatrix}
        \bm{\theta} \\ \mathbf{r}
    \end{bmatrix}
     = - G \nabla H(\bm{\theta}, \mathbf{r}) dt 
    \label{equ:sto_sys_copy}
\end{equation}

The reason we have reformulated the equations into the system given by \ref{equ:sto_sys_copy} is to demonstrate the time reversibility of the equations even with the added noise. \ref{equ:sto_sys} is the forward-time Markov process. The reverse-time Markov process is given by 

\begin{equation}
    d \begin{bmatrix}
        \bm{\theta} \\ \mathbf{-\mathbf{r}}
    \end{bmatrix}
     = G \nabla H(\bm{\theta}, \mathbf{-\mathbf{r}}) dt 
    \label{equ:sto_sys_rev}
\end{equation}

In order for equations to exhibit time reversibility, \ref{equ:sto_sys} and \ref{equ:sto_sys_rev} must be equivalent. If we expand out the position component of the reverse-time Markov process we have 

\begin{equation}
    d \bm{\theta} = - I M^{-1}(-\mathbf{r})dt = M^{-1} \mathbf{r} dt 
\end{equation}

and the momentum component

\begin{equation}
    d(-\mathbf{r}) = \nabla U(\bm{\theta})
\end{equation}

Noting that our normal distribution for the gradient noise is centered on zero and therefore symmetric. Therefore we have 

\begin{equation}
    d \begin{bmatrix}
        \bm{\theta} \\ \mathbf{r}
    \end{bmatrix}
     = 
     d \begin{bmatrix}
        \bm{\theta} \\ -\mathbf{r}
    \end{bmatrix}
\end{equation}

\subsection{Formulation of HMC Dynamics in Terms of Fokker-Planck/Continuity Equation} \label{app:fokker_planck_hmc}
In the original SGHMC paper \cite{chen2014stochastic}, the SGHMC dynamics are put in the form of a Fokker-Planck equation. We will do the same here. The Fokker Planck equation (FPE) describes a the time evolution of a probability density function under both determinist and random forces via a stochastic differential equation (SDE) in the form of 

\begin{equation}
    d\mathbf{z} = g(\mathbf{z})dt + \sqrt{2D(\mathbf{z})dt}dW
\end{equation}

where the time evolution of the distribution of $z$ ($p_t(\mathbf{z})$) can be denoted as 
\begin{equation}
    \partial_t p_t(\mathbf{z}) = - \nabla^T \left[g(\mathbf{z})p_t(\mathbf{z})\right] + \nabla^T[D(\mathbf{z})p_t(\mathbf{z})]
\end{equation}

Using the SDE form of Hamiltonian dynamics given in appendix \ref{app:time_rev_sghmc} equation \ref{equ:sto_sys}, we can put this in terms of a corresponding FPE by using $g(\mathbf{z}) = - G \nabla H(\bm{\theta}, \mathbf{r})$. 
\begin{equation}
    \partial_t p_t(\bm{\theta}, \mathbf{r}) = \nabla^T \left[G \nabla H(\bm{\theta}, \mathbf{r})p_t(\bm{\theta}, \mathbf{r})\right] + \nabla^T[D(\bm{\theta}, \mathbf{r})p_t(\bm{\theta}, \mathbf{r})]
\end{equation}

Notice how in our case, $D(\bm{\theta}, \mathbf{r}) = D = \begin{bmatrix} 0 & 0\\ 0 & 0 \end{bmatrix}$ as we do not have the stochastic term. Therefore our equation reduces to 
\begin{equation}
    \partial_t p_t(\bm{\theta}, \mathbf{r}) = \nabla^T \left[G \nabla H(\bm{\theta}, \mathbf{r})p_t(\bm{\theta}, \mathbf{r})\right] 
    \label{eq:continuity_hmc}
\end{equation}

When there is no stochastic/diffusion term, the equation simply becomes a continuity equation. 

\subsection{Constant Entropy Proof} \label{app:entropy}
Similarly to \cite{chen2014stochastic} we shall show that this HMC continuity equation \ref{eq:continuity_hmc} introduces no entropy over time by using the differential form of entropy
\begin{equation}
    h(x) = \int_{\chi} f(x)dx  
\end{equation}

Where $f(x) = x \ln x$, $\mathbf{z} = (\bm{\theta}, \mathbf{r})$ and $x = p_t(\mathbf{z})$. We have to show that the time evolution of this entropy equation is zero, i.e. we have to show 
\begin{equation}
    \partial_t h(p_t(\mathbf{z})) = -\partial_t \int_{z} f(p_t(\mathbf{z}))d\mathbf{z} = 0
    \label{eq:time_entropy}
\end{equation}

Using the chain rule $\frac{\partial}{\partial x} f(g(x)) = f'(g(x))g'(x)$, we can rewrite \ref{eq:time_entropy} in the following form 
\begin{equation}
    \partial_t h(\mathbf{z}) = -\int_z \partial_t f(p_t(\mathbf{z})) \partial_t p_t(\mathbf{z}) d\mathbf{z} 
\end{equation}

Noticing that $\partial_t p_t(\mathbf{z}) = \partial_t p_t(\bm{\theta}, \mathbf{r}) = \nabla^T \left[G \nabla H(\bm{\theta}, \mathbf{r})p_t(\bm{\theta}, \mathbf{r})\right]$
\begin{equation}
    \partial_t h(\mathbf{z}) = -\int_z  f'(p_t(\mathbf{z})) \nabla^T \left[G \nabla H(\mathbf{z})\right]p_t(\mathbf{z}) d\mathbf{z} 
\end{equation}

We can now use the product rule $\frac{\partial}{\partial x}(uv) = u\frac{\partial v}{\partial x} + v\frac{\partial u}{\partial x}$ where $u = G \nabla H(\mathbf{z})$ and $v = p_t(\mathbf{z})$
\begin{equation}
    \begin{aligned}
        \partial_t h(\mathbf{z}) &= -\int_z  f'(p_t(\mathbf{z})) \nabla^T [G \nabla H(\mathbf{z})] p_t(\mathbf{z}) d\mathbf{z} \\
        & -\int_z  f'(p_t(\mathbf{z})) \nabla^T p_t(\mathbf{z}) [G \nabla H(\mathbf{z})] d\mathbf{z} 
    \end{aligned}  
\end{equation}

\begin{equation}
    \begin{aligned}
        \nabla^T [G\nabla H(\mathbf{z})] &= 
        \begin{bmatrix}
            \frac{\partial}{\partial \bm{\theta}} & \frac{\partial}{\partial \mathbf{r}}
        \end{bmatrix} G
        \begin{bmatrix}
            \frac{\partial}{\partial \bm{\theta}} \\ \frac{\partial}{\partial \mathbf{r}}
        \end{bmatrix}
        H(\bm{\theta}, \mathbf{r}) \\
        &= [\partial_{\bm{\theta}}, \partial_{\mathbf{r}}]G
        \begin{bmatrix}
            \partial_{\theta} \\ \partial_{\mathbf{r}}
        \end{bmatrix} \\
        &= [\partial_{\bm{\theta}}, \partial_{\mathbf{r}}]
        \begin{bmatrix}
            0 & -I \\ I & 0
        \end{bmatrix}
        \begin{bmatrix}
            \partial_{\theta} \\ \partial_{\mathbf{r}}
        \end{bmatrix} \\
        &= \nabla^T
        \begin{bmatrix}
            \partial_{\theta} H \\ \partial_{\mathbf{r}} H
        \end{bmatrix} \\
        &= - \partial_{\bm{\theta}} \partial_{\mathbf{r}} H + \partial_{\mathbf{r}} \partial_{\bm{\theta}} H
    \end{aligned}
    \label{eq:sep_ham}
\end{equation}

As the Hamiltonian is separable, the differentials are commutative and therefore the expression given by \ref{eq:sep_ham} is equal to 0 and therefore
\begin{equation}
    \begin{aligned}
        \partial_t h(\mathbf{z}) &=  -\int_z  f'(p_t(\mathbf{z})) \nabla^T p_t(\mathbf{z}) [G \nabla H(\mathbf{z})] d\mathbf{z} \\
        &= -\int_z  f'(p_t(\mathbf{z})) (\nabla p_t(\mathbf{z}))^T [G \nabla H(\mathbf{z})] d\mathbf{z}
    \end{aligned}
\end{equation}

The expression $f'(p_t(\mathbf{z})) (\nabla p_t(\mathbf{z}))$ can be rearranged via the chain rule $f'(g(x))g'(x) = \frac{\partial}{\partial x} f(g(x))$ and therefore
\begin{equation}
    \begin{aligned}
        \partial_t h(\mathbf{z}) &=  -\int_z (\nabla f(p_t(\mathbf{z})))^T [G \nabla H(\mathbf{z})] d\mathbf{z} 
    \end{aligned}
\end{equation}

The using integration by parts
\begin{equation}
    \begin{aligned}
        \partial_t h(\mathbf{z}) &=  -\left[  f(p_t(\mathbf{z}))  [G \nabla H(\mathbf{z})] \right] \Big|_z \\
        &= -\int_z  f(p_t(\mathbf{z})) \nabla^T [G \nabla H(\mathbf{z})] d\mathbf{z}
    \end{aligned}
\end{equation}

We already know that $\nabla^T [G \nabla H(\mathbf{z})]$ is equal to 0. 
\begin{equation}
    \begin{aligned}
        \partial_t h(\mathbf{z}) =  -\left[  f(p_t(\mathbf{z}))  [G \nabla H(\mathbf{z})] \right] \Big|_z 
    \end{aligned}
\end{equation}

As $f(x) = x\ln x$, our function is defined as $f(p_t(\mathbf{z})) \ln p_t(\mathbf{z})$ which is defined on the interval $(0, \infty)$. as $x \xrightarrow[]{} 0$, $f(x) \xrightarrow[]{} 0$. As $p_t(\mathbf{z})$ is a probability density function, it must go to zero at $\infty$ to be normalizable, a condition of a valid probability distribution. Therefore, $z \xrightarrow[]{} \infty$, $f(p_t(\mathbf{z}))  [G \nabla H(\mathbf{z})] \xrightarrow[]{} 0$. 

\newpage

\section{Mini-Batch Estimate is Unbiased} \label{app:unbiased_grads}

The following is a proof that the approximation of our full-batch gradient is an unbiased estimate and can be interpreted as an extension of the mini-batch gradient descent bias proof. This result has been shown before but we provide it in full for completeness sake. 

Our full-batch gradient estimate can be written as: 
\begin{equation}
    \nabla \Tilde{U}(\Tilde{\mathcal{D}} | \bm{\theta}) = - \frac{|\mathcal{D}|}{|\Tilde{\mathcal{D}}|} \sum_{\mathbf{x} \in \Tilde{\mathcal{D}}} \nabla \log p(\mathbf{x} | \bm{\theta}) - \nabla \log p(\bm{\theta}).
\end{equation}

Where $\mathcal{D}$ is the full dataset and $\Tilde{\mathcal{D}}$ is the mini-batch. In order to prove that this estimate is unbiased, we need to show 
\begin{equation}
    bias = \mathbb{E}\left[\nabla \Tilde{U}(\Tilde{\mathcal{D}} | \bm{\theta})\right] - \nabla U(\mathcal{D} | \bm{\theta}),
\end{equation}

\begin{equation}
    \nabla U(\mathcal{D} | \bm{\theta}) = - \sum_{\mathbf{x} \in \mathcal{D}} \nabla \log p(\mathbf{x} | \bm{\theta}) - \nabla \log p(\bm{\theta}).
    \label{eq:full_grad}
\end{equation}

First we take the expectation of the approximation 
\begin{equation}
    \mathbb{E}\left[\nabla \Tilde{U}(\Tilde{\mathcal{D}} | \bm{\theta})\right] = \mathbb{E}\left[- \frac{|\mathcal{D}|}{|\Tilde{\mathcal{D}}|} \sum_{\mathbf{x} \in \Tilde{D}} \nabla \log p(\mathbf{x} | \bm{\theta}) - \nabla \log p(\bm{\theta})\right].
\end{equation}

Using the linearity of expectations 
\begin{equation}
    \mathbb{E}\left[\nabla \Tilde{U}(\Tilde{\mathcal{D}} | \bm{\theta})\right] = - \frac{|\mathcal{D}|}{|\Tilde{\mathcal{D}}|} \mathbb{E}\left[\sum_{\mathbf{x} \in \Tilde{\mathcal{D}}} \nabla \log p(\mathbf{x} | \bm{\theta})\right] - \nabla \log p(\bm{\theta}).
    \label{eq:expect_grad_part_1}
\end{equation}

Assuming the samples in the mini-batch are uniformly randomly sampled without replacement
\begin{equation}
    \mathbb{E}\left[\sum_{\mathbf{x} \in \Tilde{\mathcal{D}}} \nabla \log p(\mathbf{x} | \bm{\theta})\right] = \frac{1}{|\mathcal{D}|}\sum_{\mathbf{x} \in \mathcal{D}} P(x \in \mathcal{D})\nabla \log p(\mathbf{x} | \bm{\theta}),
\end{equation}

\begin{equation}
    \mathbb{E}\left[\sum_{\mathbf{x} \in \Tilde{\mathcal{D}}} \nabla \log p(\mathbf{x} | \bm{\theta})\right] = \frac{|\Tilde{\mathcal{D}}|}{|\mathcal{D}|}\sum_{\mathbf{x} \in \mathcal{D}} \nabla \log p(\mathbf{x} | \bm{\theta}),
    \label{eq:expect_grad_part_2}
\end{equation}

We can use the result from \ref{eq:expect_grad_part_2} and sub it into \ref{eq:expect_grad_part_1} giving us 

\begin{equation}
    \mathbb{E}\left[\nabla \Tilde{U}(\Tilde{\mathcal{D}} | \bm{\theta})\right] = - \frac{|\mathcal{D}|}{|\Tilde{\mathcal{D}}|} \cdot \frac{|\Tilde{\mathcal{D}}|}{|\mathcal{D}|} \sum_{\mathbf{x} \in \mathcal{D}} \nabla \log p(\mathbf{x} | \bm{\theta}) - \nabla \log p(\bm{\theta}),
\end{equation}

which simplifies to 
\begin{equation}
    \mathbb{E}\left[\nabla \Tilde{U}(\Tilde{\mathcal{D}} | \bm{\theta})\right] = - \sum_{\mathbf{x} \in \mathcal{D}} \nabla \log p(\mathbf{x} | \bm{\theta}) - \nabla \log p(\bm{\theta}).
    \label{eq:expect_grad_part_3}
\end{equation}

We notice that \ref{eq:expect_grad_part_3} is the full-batch gradient evaluation given by \ref{eq:full_grad}. We have therefore shown that 
\begin{equation}
    \mathbb{E}\left[\nabla \Tilde{U}(\Tilde{\mathcal{D}} | \bm{\theta})\right] = \nabla U(\mathcal{D} | \bm{\theta}),
\end{equation}

and therefore 
\begin{equation}
    bias = \mathbb{E}\left[\nabla \Tilde{U}(\Tilde{\mathcal{D}} | \bm{\theta})\right] - \nabla U(\mathcal{D} | \bm{\theta}) = 0,
\end{equation}

which shows that this gradient approximation we use is unbiased. 

\section{Unlearning Overfitting} \label{app:unlearn_overfit}

During our experiments, we observed that at the start of the SMC sampling phase, the training loss of the models initially increased, while the validation loss remained stable. This suggests that the networks had overfitted during the SGD training phase: the models had minimized the training loss excessively without corresponding improvements on the validation set.

Interestingly, the sampling procedure not only improved final test metrics, but also appeared to \emph{unlearn} some of the overfitting. As the sampling progressed, the models moved away from sharp local optima that fit the training data well but failed to generalize.

\begin{figure}[H]
    \centering
    \begin{minipage}{0.45\textwidth}
        \centering
        \includegraphics[width=\textwidth]{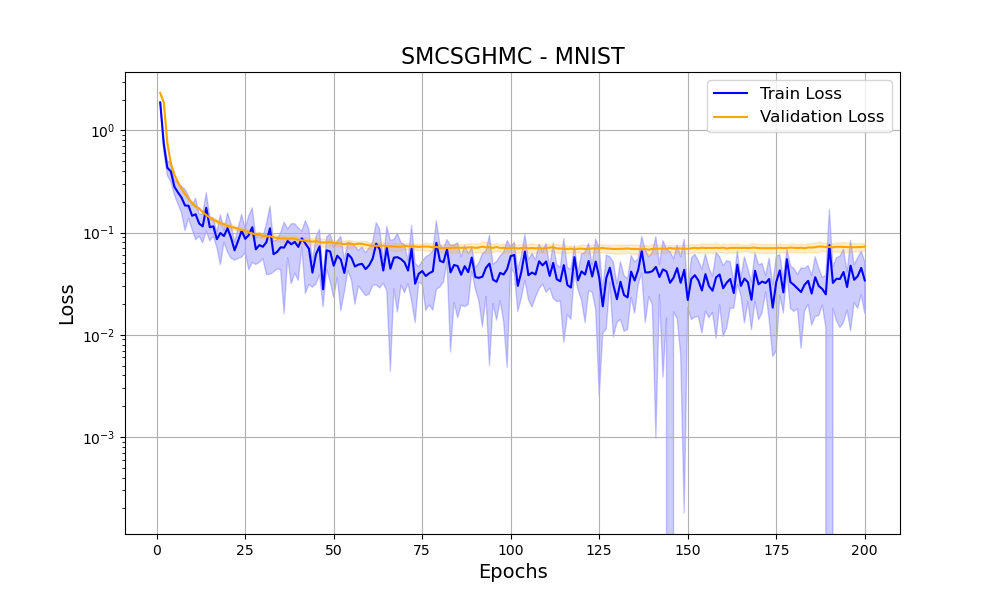}
        \caption{Training and validation loss for MNIST.}
        \label{fig:train_smc_mnist}
    \end{minipage}
    \hfill
    \begin{minipage}{0.45\textwidth}
        \centering
        \includegraphics[width=\textwidth]{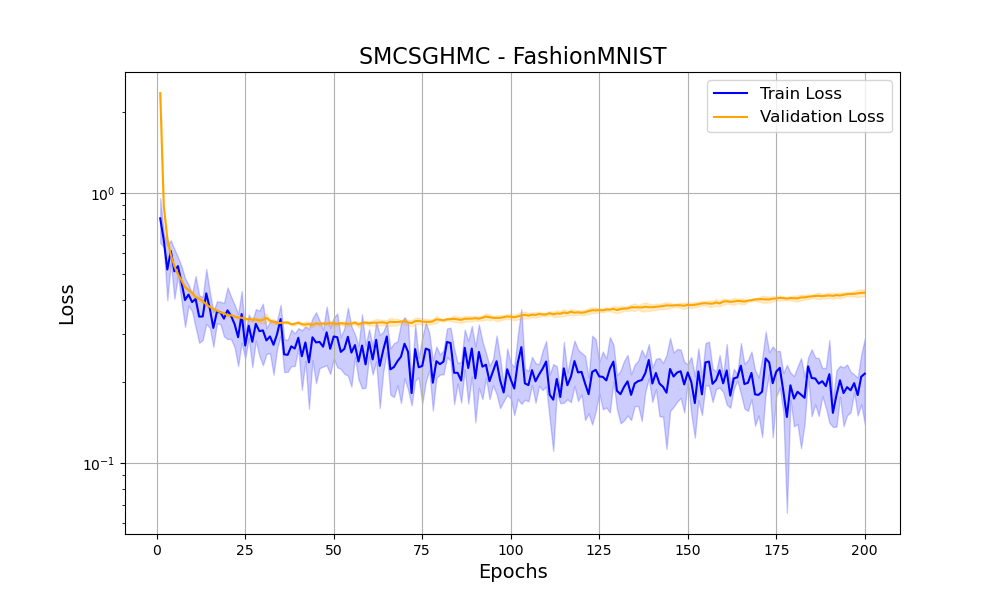}
        \caption{Training and validation loss for FashionMNIST.}
        \label{fig:train_smc_fashion}
    \end{minipage}
    
    \vspace{0.5cm} 
    
    \begin{minipage}{0.45\textwidth}
        \centering
        \includegraphics[width=\textwidth]{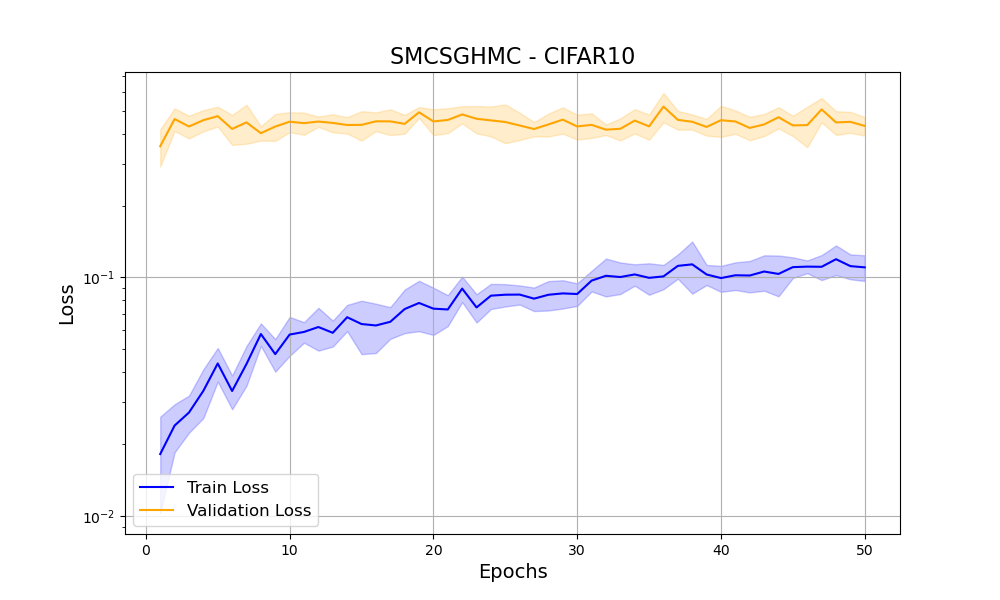}
        \caption{Training and validation loss for CIFAR-10.}
        \label{fig:train_smc_cifar_repeat}
    \end{minipage}
    \hfill
    \begin{minipage}{0.45\textwidth}
        \centering
        \includegraphics[width=\textwidth]{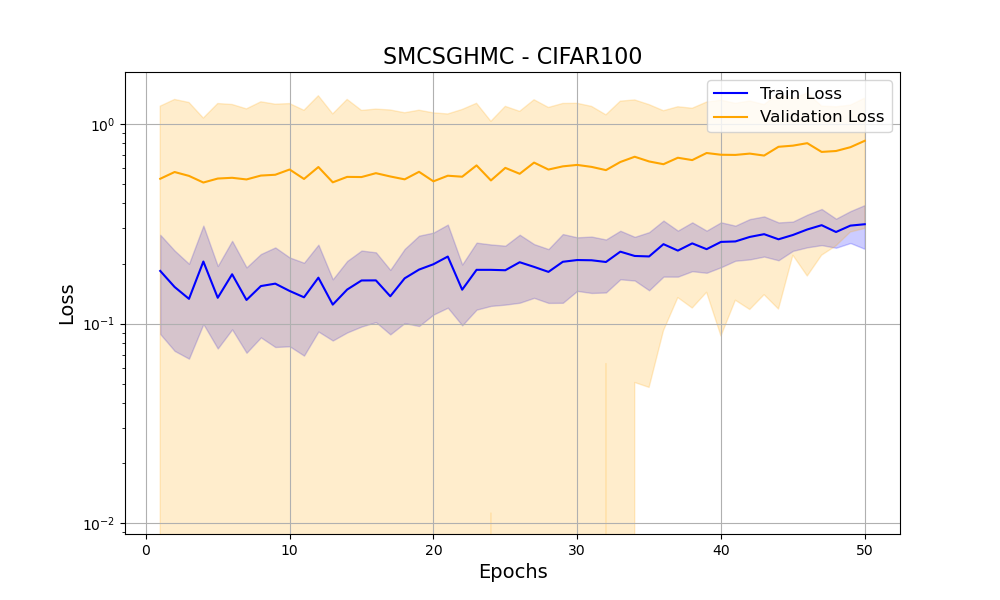}
        \caption{Training and validation loss for CIFAR-100.}
        \label{fig:train_smc_cifar100_repeat}
    \end{minipage}
\end{figure}

Figures~\ref{fig:train_smc_mnist}--\ref{fig:train_smc_cifar100_repeat} illustrate the training and validation loss trajectories across different datasets.

In standard deep learning, early stopping (ES)~\cite{bai2021understandingimprovingearlystopping} is a common technique to prevent overfitting by monitoring validation loss. Typically, overfitting is diagnosed when the validation loss starts increasing while the training loss continues decreasing. 
In contrast, during our sampling phase, the opposite occurs: the training loss increases while validation performance remains stable or improves, indicating that the sampling procedure smooths over spurious overfit solutions. This result can be more prominently seen in \ref{fig:train_smc_cifar_repeat}. It can also be seen in the beginning of the training phase in \ref{fig:train_smc_cifar100_repeat} however, we also notice the slight increase in the validation loss in the latter part of the training phase as discussed in \ref{sec:further_work}. 

\vspace{0.5cm} 

\vfill

\section{Supplementary Experiment Details} \label{app:supplement_experi}
\subsection{Experimental Set Up Parameters} \label{app:experi_setup}

For all experiments, the results were generated using an Nvidia A100 GPU 80Gb. 

For the synthetic dataset we used $N = 1000$ samples for $K = 400$ iterations. The first $200$ iterations were treated as warm up samples. A step size of $h = 0.2$ and a trajectory length of $L = 10$ were used for the HMC proposal.

For the MNIST and FashionMNIST datasets, networks were trained from scratch using an SMC sampler. Specifically, LeNet \cite{lecun1998mnist} was used for MNIST, and a two-layer convolutional network was used for FashionMNIST. The hyperparameters for these experiments were set to $N = 10$, $K = 200$, and $h = 2 \times 10^{-5}$.

However, we observed that training larger and more complex architectures from scratch was computationally slow. As a result, for the larger experiments, we applied our method to pretrained networks and used these models for comparisons with other algorithms. The MNIST and FashionMNIST results are included for completeness.

The data splits for each experiment are as follows: 
\begin{itemize}
    \item \textbf{MNIST/FashionMNIST:} Train/Validation/Test - 48000/12000/10000
    \item \textbf{CIFAR10/CIFAR100:} Train/Validation/Test - 45000/5000/10000
\end{itemize}

Table \ref{tab:hyper_parameters} gives the overview of the experiment parameters for the CIFAR classification and OOD tasks. 

\begin{table}[H]
    \centering
    \caption{Hyperparameter configurations for different methods and experiments.}
    \label{tab:hyper_parameters}
    \renewcommand{\arraystretch}{1.15}
    \resizebox{0.6\textwidth}{!}{
    \begin{tabular}{l@{\hskip 3pt}c@{\hskip 3pt}c@{\hskip 3pt}c@{\hskip 3pt}c@{\hskip 3pt}c@{\hskip 3pt}c@{\hskip 3pt}c@{\hskip 3pt}c}
        \toprule
        \textbf{Method} & \textbf{Batch Size} & \textbf{$N$} & \textbf{$h$} & \textbf{$K$} & \textbf{$M$} & \textbf{$\nu$} & \textbf{Anneal $h$} & \textbf{Samples} \\
        \midrule
        \multicolumn{9}{c}{\textbf{CIFAR-10: ResNet20-FRN}} \\
        \midrule
        SMC       & 500 & 10 & 2e-5  & 50  & 25 & N/A  & No  & 250 \\
        SGD       & 128 & 1  & 0.01  & 200 & -- & 1e-2 & Yes & -- \\
        Ensemble  & 128 & 5  & 0.01  & 200 & -- & 1e-2 & Yes & -- \\
        SWAG Diag & 128 & 1  & 0.001 & --  & -- & --   & No  & 30 \\
        SWAG      & 128 & 1  & 0.001 & --  & -- & --   & No  & 30 \\
        \midrule
        \multicolumn{9}{c}{\textbf{CIFAR-100: VGG16}} \\
        \midrule
        SMC       & 500 & 5  & 2e-6  & 50  & 25 & N/A  & No  & 125 \\
        SGD       & 256 & 1  & 0.01  & 200 & -- & 5e-4 & Yes & -- \\
        Ensemble  & 256 & 5  & 0.01  & 200 & -- & 5e-4 & Yes & -- \\
        SWAG Diag & 256 & 1  & 0.001 & --  & -- & --   & No  & 30 \\
        SWAG      & 256 & 1  & 0.001 & --  & -- & --   & No  & 30 \\
        \midrule
        \multicolumn{9}{c}{\textbf{OOD: VGG16}} \\
        \midrule
        SMC       & 500 & 5  & 2e-6  & 50  & 25 & N/A  & No  & 125 \\
        SGD       & 256 & 1  & 0.01  & 200 & -- & 5e-4 & Yes & -- \\
        Ensemble  & 256 & 5  & 0.01  & 200 & -- & 5e-4 & Yes & -- \\
        SWAG Diag & 256 & 1  & 0.001 & --  & -- & --   & No  & 30 \\
        SWAG      & 256 & 1  & 0.001 & --  & -- & --   & No  & 30 \\
        \bottomrule
    \end{tabular}
    }
\end{table}

\subsection{Energy Based Scoring for OOD Detection} \label{app:energy_scores}

Energy-based scoring \cite{liu2021energybasedoutofdistributiondetection} is a method for Out-of-Distribution (OOD) detection that leverages the output logits of a neural network to compute a scalar energy score. The energy function is inspired by energy-based models (EBMs), where lower energy scores correspond to higher confidence that an input belongs to the in-distribution (ID) data, and higher energy scores suggest the input might be OOD.

Given a neural network with output logits $\mathbf{z} = [z_1, z_2, \dots, z_C]$, where $z_i$ is the logit for the $i$-th class, the energy score $E(\mathbf{z})$ is defined as:
\begin{equation}
E(\mathbf{z}) = -\log \sum_{i=1}^{C} e^{z_i},
\end{equation}
where $C$ is the number of classes in the ID dataset.

This energy formulation is directly related to the log-partition function in energy-based models and avoids the need for softmax normalization, making it robust to overconfidence issues often encountered in OOD scenarios. During OOD detection, the energy score is used to classify inputs into either ID or OOD categories. A threshold $\tau$ is determined on the energy scores, typically using a validation set that includes known OOD samples. The classification rule is given by:
\[
\text{Label} =
\begin{cases} 
\text{ID}, & \text{if } E(\mathbf{z}) \leq \tau, \\
\text{OOD}, & \text{if } E(\mathbf{z}) > \tau.
\end{cases}
\]

To further enhance OOD detection, models can be fine-tuned with energy-based regularization. For instance:
\begin{itemize}
    \item \textbf{Energy-Regularized Learning}: Introduces a penalty in the loss function to encourage higher energy for OOD inputs.
    \item \textbf{Temperature Scaling}: Adjusts the sensitivity of logits by scaling with a temperature parameter $T$:
    \begin{equation}
    E_T(\mathbf{z}) = -\log \sum_{i=1}^{C} e^{z_i / T}.
    \end{equation}
\end{itemize}

In our tests, we have not done any fine tuning of the network such as temperature adjustment or loss function penalties. In order to calculate the energy threshold, we use the TPR95 criteria, that is we use the energy level which classifies 95\% of the in distribution data samples correctly.

\subsection{AUROC Curves} \label{app:auroc_curves}

\begin{figure}[H]
    \centering
    \begin{minipage}{0.45\textwidth}
        \centering
        \includegraphics[width=1.0\textwidth]{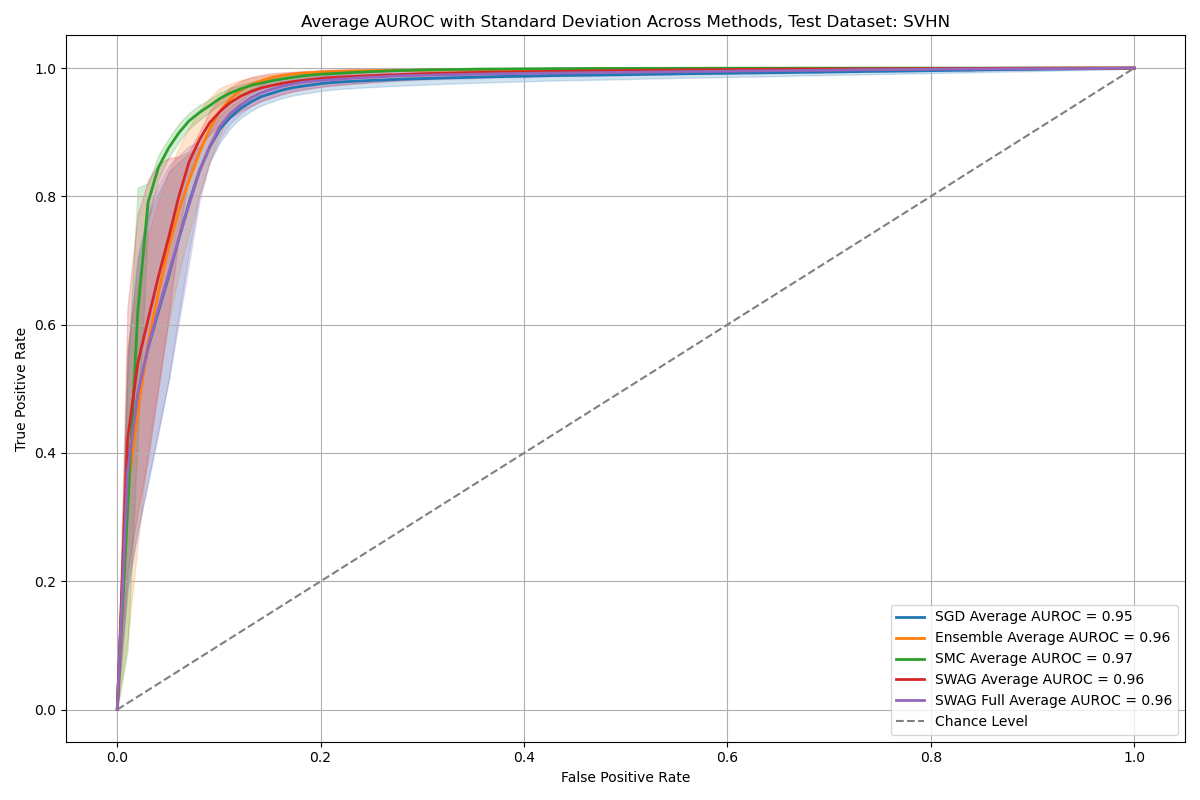} 
        \caption{AUROC comparison for SVHN OOD data.}
        \label{fig:auroc_svhn}
    \end{minipage}
    \hfill
        \begin{minipage}{0.45\textwidth}
        \centering
        \includegraphics[width=1.0\textwidth]{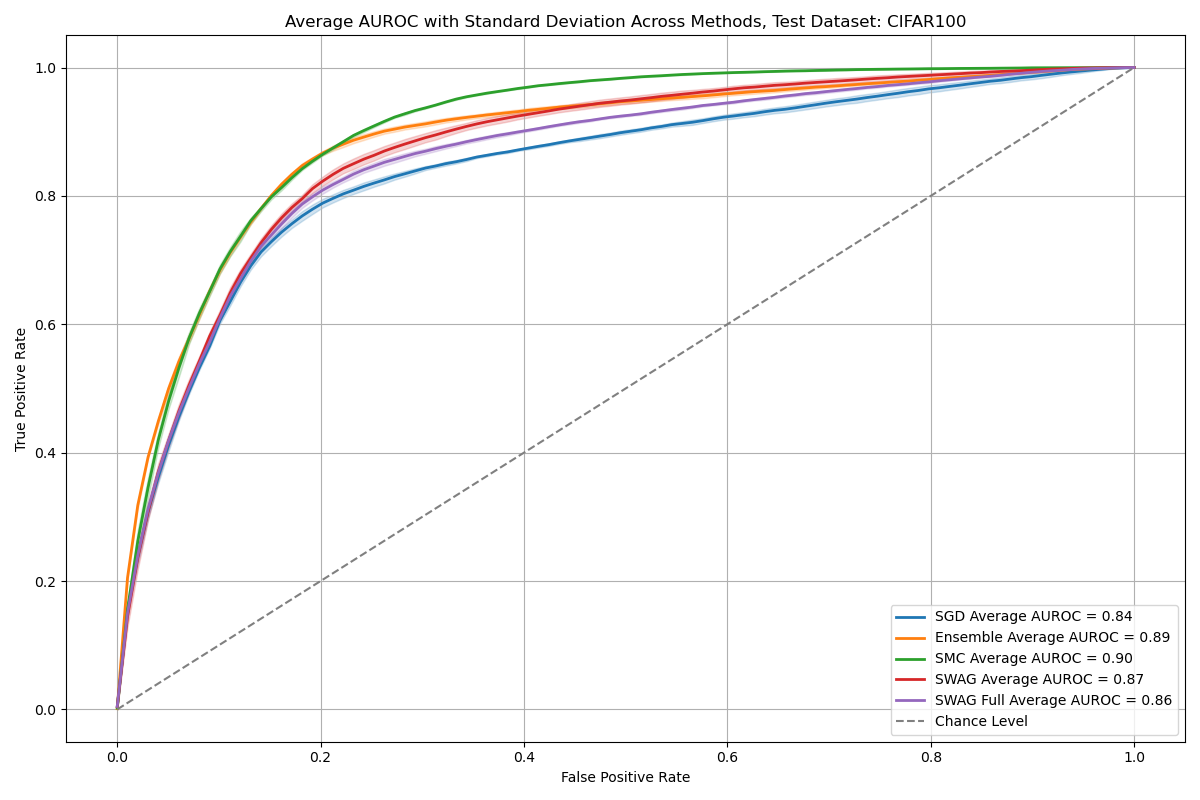} 
        \caption{AUROC comparison for CIFAR100 OOD data.}
        \label{fig:auroc_cifar100}
    \end{minipage}
    \vspace{1cm}
    \begin{minipage}{0.45\textwidth}
        \centering
        \includegraphics[width=1.0\textwidth]{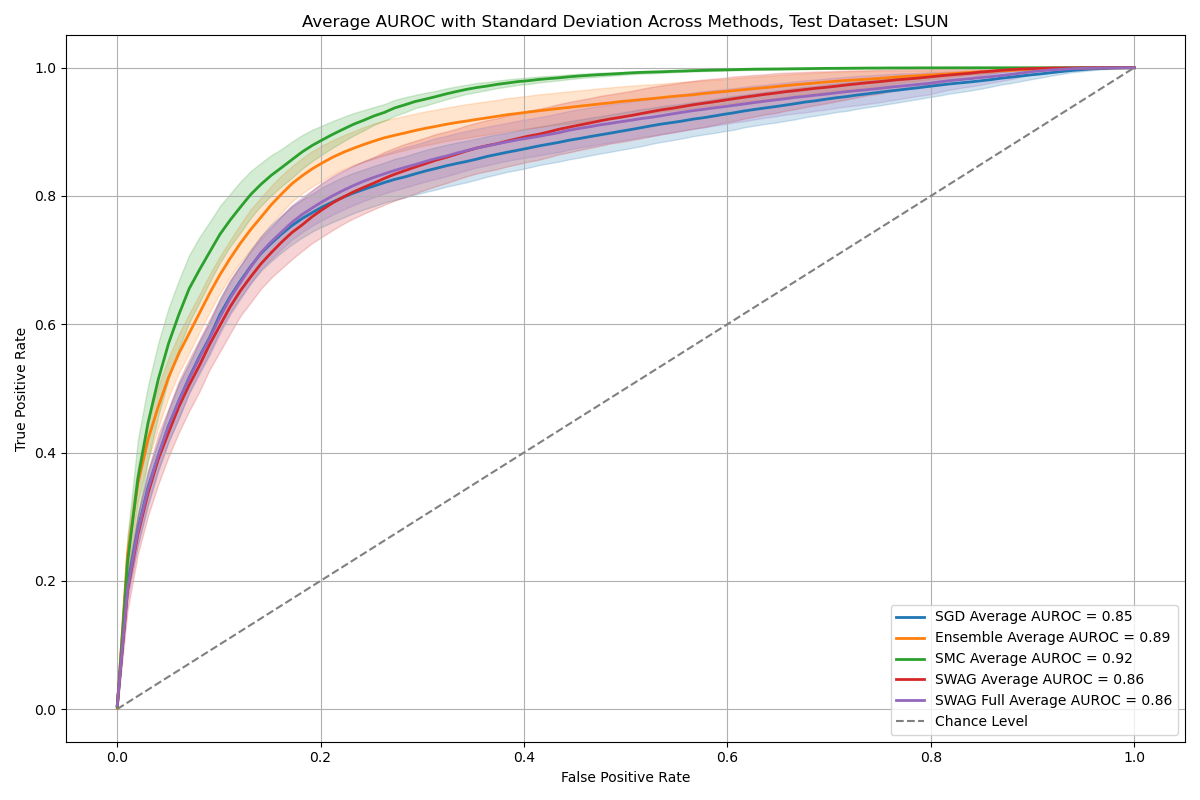} 
        \caption{AUROC comparison for LSUN church OOD data.}
        \label{fig:auroc_lsun_church}
    \end{minipage}
    \hfill
        \begin{minipage}{0.45\textwidth}
        \centering
        \includegraphics[width=1.0\textwidth]{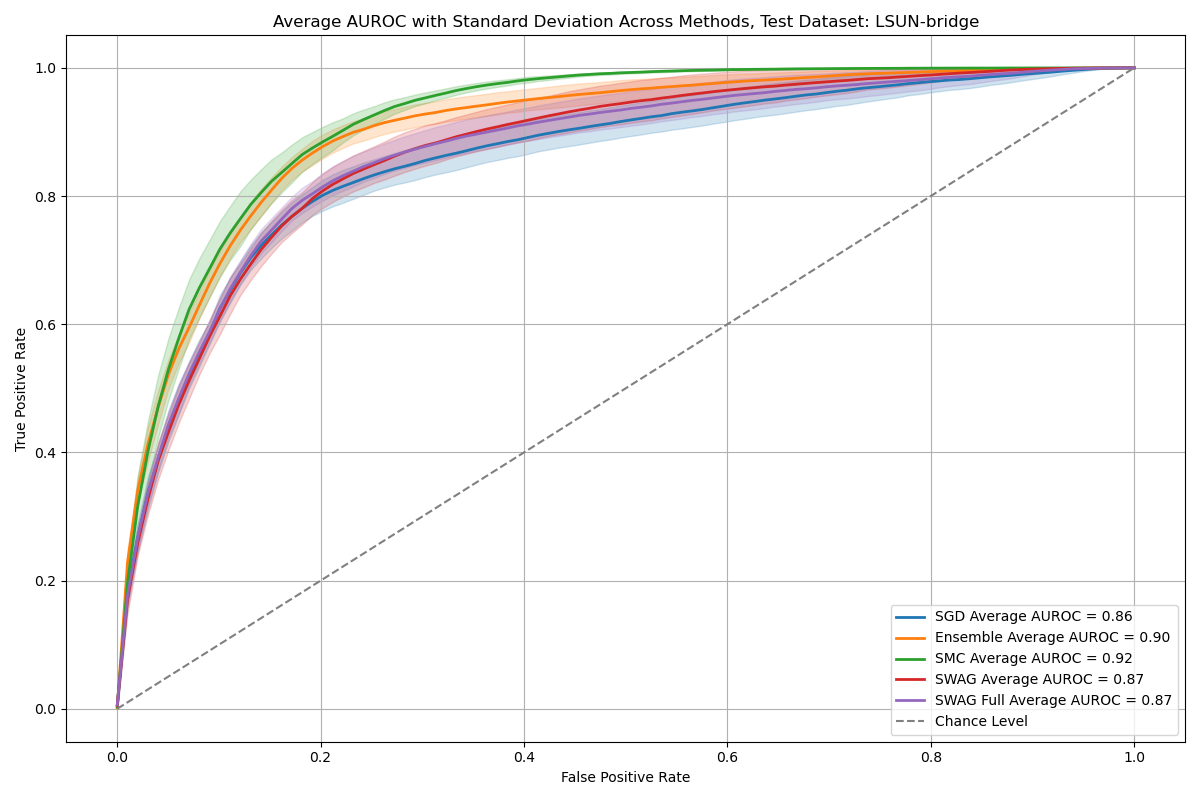} 
        \caption{AUROC comparison for LSUN bridge OOD data.}
        \label{fig:auroc_lsun_bridge}
    \end{minipage}
\end{figure}

\vfill 
\newpage

\section{Weighted Ensemble for SMC trained BNNs} \label{app:weighted_ensemble}
Weighted ensembles are a widely used technique in machine learning that combine multiple individual models to form a stronger predictive system. Each model contributes to the final prediction with an influence proportional to its assigned weight. In this work, the models and their corresponding weights are obtained by the Sequential Monte Carlo (SMC) sampling procedure. During training, the sampler generates a collection of model parameters \( \{\bm{\theta}_1, \bm{\theta}_2, \ldots, \bm{\theta}_N\} \) along with associated (unnormalized) importance weights \( \{\mathbf{w}_1, \mathbf{w}_2, \ldots, \mathbf{w}_J\} \).

After the sampling phase, the weights are normalized to sum to one:

\begin{equation}
    \Tilde{\mathbf{w}}_j = \frac{\mathbf{w}_j}{\sum_{j=1}^{J} \mathbf{w}_j} \quad \text{for each} \quad j = 1, \ldots, J,
\end{equation}

ensuring that:

\begin{equation}
    \sum_{j=1}^{J} \Tilde{\mathbf{w}}_j = 1, \quad \text{and} \quad \mathbf{w}_j \geq 0 \quad \forall j.
\end{equation}

The final ensemble prediction aggregates the outputs of all sampled models, weighted according to these normalized importance weights:

\begin{equation}
    \hat{\mathbf{y}}_{\text{pred}} = \arg\max_k \left( \sum_{j=1}^{J} \Tilde{\mathbf{w}}_j \, P(f(\mathbf{c}_k \mid \mathbf{x}, \bm{\theta}_j)) \right).
\end{equation}

where:
\begin{itemize}
    \item \( P(\mathbf{c}_k \mid \mathbf{x}, \bm{\theta}_j) \) is the probability assigned to class \( \mathbf{c}_k \) by the neural network model \( f \) for input \( \mathbf{x} \),
    \item \( \arg\max_k \) selects the class with the highest weighted sum of probabilities.
\end{itemize}

This approach allows the ensemble to fully leverage the posterior distribution approximated by the SMC procedure, capturing model uncertainty and diversity. 

\vfill
\newpage

\section{Further Results}

\subsection{Number of Samples Comparison between SMC and SWAG.} \label{app:swag_comp_samples}
This section explores the differences the number of collected samples has on the CIFAR10/100 datasets for both the SWAG methods and SMC. Table \ref{tab:samples_comp} gives the results on the CIFAR10 dataset for a varying number of samples. We see that the results are similar to the ones reported in table \ref{tab:image_results} in that the ECE and loss are consistently better despite the reduced number of samples. Even with the same number of samples as we do models in the ensembles ($N=5$), the ECE score for SMC is better. The tables also show that as we increase the number of samples, the metrics for SMC improve for every metric, while the SWAG methods performance is steady for different sample numbers. 

\begin{table}[H]
    \centering
    \caption{CIFAR-10/100 performance with different number of samples for Swag and SMC}
    \label{tab:samples_comp}
    \renewcommand{\arraystretch}{1.2}
    \resizebox{\textwidth}{!}{
    \begin{tabular}{lcccccc}
        \toprule
        \multicolumn{1}{c}{} & \multicolumn{3}{c}{\textbf{CIFAR10}} & \multicolumn{3}{c}{\textbf{CIFAR100}} \\
        \hline
        \textbf{Method} & \textbf{Accuracy} & \textbf{Loss} & \textbf{ECE} & \textbf{Accuracy} & \textbf{Loss} & \textbf{ECE} \\
        \midrule
        \multicolumn{7}{c}{\textbf{N = 5}} \\
        \midrule
        SMC       & 90.53 ± 0.48 & \textbf{0.3439 ± 0.0166} & \textbf{0.0542 ± 0.0025} & 70.31 ± 0.70 & \textbf{1.2804 ± 0.0536} & \textbf{0.1089 ± 0.0095} \\
        SWAG Diag & 90.71 ± 0.34 & 0.4218 ± 0.0076 & 0.0645 ± 0.0015 & 71.54 ± 0.28 & 1.7474 ± 0.0415 & 0.1938 ± 0.0040  \\
        SWAG      & \textbf{90.72 ± 0.30} & 0.4269 ± 0.0087 & 0.0654 ± 0.0019 & \textbf{71.58 ± 0.38} & 1.7293 ± 0.0564 & 0.1924 ± 0.0042  \\
        \midrule
        \multicolumn{7}{c}{\textbf{N = 10}} \\
        \midrule
        SMC       & \textbf{90.84 ± 0.30} & \textbf{0.3257 ± 0.0084} & \textbf{0.0521 ± 0.0025} & 70.80 ± 0.25 & \textbf{1.2439 ± 0.0190} & \textbf{0.1033 ± 0.0064} \\
        SWAG Diag & 90.73 ± 0.34 & 0.4215 ± 0.0084 & 0.0647 ± 0.0014 & 71.54 ± 0.28 & 1.7476 ± 0.0428 & 0.1933 ± 0.0042 \\
        SWAG      & 90.72 ± 0.30 & 0.4258 ± 0.0098 & 0.0652 ± 0.0018 & \textbf{71.58 ± 0.26} & 1.7413 ± 0.0276 & 0.1934 ± 0.0023 \\
        \midrule
        \multicolumn{7}{c}{\textbf{N = 20}} \\
        \midrule
        SMC       & \textbf{91.11 ± 0.29} & \textbf{0.3133 ± 0.0082} & \textbf{0.0506 ± 0.0015} & 71.25 ± 0.18 & \textbf{1.2100 ± 0.0125} & \textbf{0.0987 ± 0.0029} \\
        SWAG Diag & 90.74 ± 0.35 & 0.4216 ± 0.0083 & 0.0646 ± 0.0017 & 71.53 ± 0.27 & 1.7474 ± 0.0421 & 0.1934 ± 0.0038 \\
        SWAG      & 90.70 ± 0.28 & 0.4260 ± 0.0069 & 0.0649 ± 0.0017 & \textbf{71.63 ± 0.45} & 1.7244 ± 0.0665 & 0.1913 ± 0.0063 \\
        \midrule
        \multicolumn{7}{c}{\textbf{N = 30}} \\
        \midrule
        SMC       & \textbf{91.09 ± 0.31} & \textbf{0.3132 ± 0.0077} & \textbf{0.0506 ± 0.0017} & 71.04 ± 0.28 & \textbf{1.2237 ± 0.0222} & \textbf{0.1007 ± 0.0077} \\
        SWAG Diag & 90.74 ± 0.34 & 0.4215 ± 0.0081 & 0.0644 ± 0.0017 & 71.53 ± 0.27 & 1.7475 ± 0.0418 & 0.1936 ± 0.0036 \\
        SWAG      & 90.70 ± 0.31 & 0.4255 ± 0.0078 & 0.0652 ± 0.0015 & \textbf{71.62 ± 0.38} & 1.7298 ± 0.0487 & 0.1921 ± 0.0041 \\
        \bottomrule
    \end{tabular}}
\end{table}

\subsection{Untempering SMC Results} \label{app:untempered_smc}
Table \ref{tab:smc_untempered_results} gives the results for the untempered SMC sampler on the CIFAR10/100 datasets. We notice that the ECE score is still better but that the loss and accuracy significantly decrease, showing the need for a tempered likelihood to produce the best results. 

\begin{table}[H]
    \centering
    \caption{Performance metrics for Untempered SMC.}
    \label{tab:smc_untempered_results}
    \begin{tabular}{lccc}
        \toprule
        \textbf{Dataset} & \textbf{Accuracy} & \textbf{Loss} & \textbf{ECE} \\
        \midrule
        CIFAR10  & 88.57 (0.52) \% & 0.4396 (0.0141) & 0.0698 (0.0016) \\
        CIFAR100 & 67.62 (0.38) \% & 1.4997 (0.0464) & 0.1359 (0.0088) \\
        \bottomrule
    \end{tabular}
\end{table}

\vfill
\newpage

\end{document}